\documentclass[10pt,twocolumn,letterpaper]{article}
\usepackage[accsupp]{axessibility} 
\usepackage[pagenumbers]{cvpr} 

\usepackage{graphicx}
\usepackage{amsmath}
\usepackage{amssymb}
\usepackage{booktabs}
\usepackage{CJKutf8}  
\usepackage{array}
\usepackage{tabularx}
\usepackage{multirow}
\usepackage{subcaption}
\usepackage[dvipsnames]{xcolor}
\usepackage{comment}
\usepackage{url}
\usepackage{colortbl}
\usepackage{nccmath}
\usepackage{makecell}
\usepackage{times}
\usepackage{epsfig}
\usepackage[pagebackref,breaklinks,colorlinks]{hyperref}

\DeclareMathAlphabet\mathbfcal{OMS}{cmsy}{b}{n}

\usepackage[capitalize]{cleveref}
\crefname{section}{Sec.}{Secs.}
\Crefname{section}{Section}{Sections}
\Crefname{table}{Table}{Tables}
\crefname{table}{Tab.}{Tabs.}

\newcolumntype{Y}{>{\centering\arraybackslash}X}

\begin{document}
\begin{CJK}{UTF8}{}
\CJKfamily{mj}

\title{Differentiable Inverse Rendering with Interpretable Basis BRDFs}

\author{
Hoon-Gyu Chung ~ ~ ~
Seokjun Choi ~ ~ ~
Seung-Hwan Baek \\
POSTECH\\
}

\twocolumn[{
\renewcommand\twocolumn[1][]{#1}
\maketitle
\begin{center}
     \vspace{-7mm}
    \centering
    \captionsetup{type=figure}
    \includegraphics[width=\linewidth]{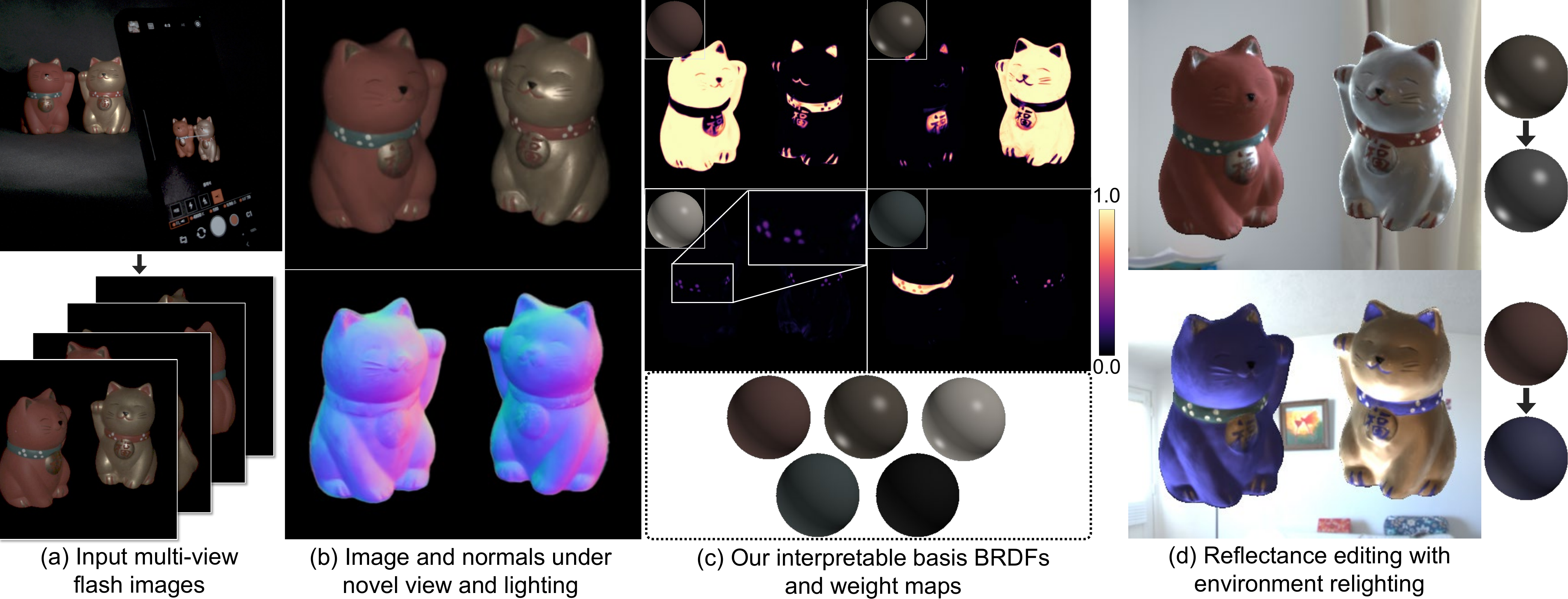}
     \vspace{-5mm}
    \captionof{figure}{We propose a differentiable inverse rendering method with (a) multi-view flash photography inputs. Our analysis-by-synthesis method achieves not only (b) novel-view relighting and accurate geometry reconstruction, but also (c) interpretable basis BRDFs and their spatially-separated weights. This allows for (d) intuitive scene editing.}
    \label{fig:teaser}
\end{center}
}]

\begin{abstract}
Inverse rendering seeks to reconstruct both geometry and spatially varying BRDFs (SVBRDFs) from captured images. To address the inherent ill-posedness of inverse rendering, basis BRDF representations are commonly used, modeling SVBRDFs as spatially varying blends of a set of basis BRDFs. However, existing methods often yield basis BRDFs that lack intuitive separation and have limited scalability to scenes of varying complexity.
In this paper, we introduce a differentiable inverse rendering method that produces interpretable basis BRDFs. Our approach models a scene using 2D Gaussians, where the reflectance of each Gaussian is defined by a weighted blend of basis BRDFs.
We efficiently render an image from the 2D Gaussians and basis BRDFs using differentiable rasterization and impose a rendering loss with the input images.
During this analysis-by-synthesis optimization process of differentiable inverse rendering, we dynamically adjust the number of basis BRDFs to fit the target scene while encouraging sparsity in the basis weights. This ensures that the reflectance of each Gaussian is represented by only a few basis BRDFs.
This approach enables the reconstruction of accurate geometry and interpretable basis BRDFs that are spatially separated. Consequently, the resulting scene representation, comprising basis BRDFs and 2D Gaussians, supports physically-based novel-view relighting and intuitive scene editing.
\end{abstract}

\section{Introduction}
\label{sec:intro}
Inverse rendering aims to reconstruct geometry and reflectance from captured images, a fundamental problem in computer vision and computer graphics. Recent advances in differentiable rendering methods~\cite{nerf, 3dgaussian} have facilitated analysis-by-synthesis differentiable inverse rendering. These methods enable the optimization of scene parameters by minimizing the discrepancy between captured and synthesized images~\cite{nerfactor, chung2024differentiable, iron, gao2023relightable, bi2024rgs}.
However, achieving accurate inverse rendering remains challenging, especially when the available lighting and viewing angles are limited. A promising approach to mitigate this problem is to exploit the spatial coherence of SVBRDFs using basis BRDF representations~\cite{matusik2003data, matusik2003efficient, trees, ren2011pocket, alldrin2008photometric}. Representing the BRDF at each point as a spatially varying blend of basis BRDFs allows gathering multiple points to accurately fit basis BRDFs and their per-point weights.

Early methods often focus solely on estimating basis BRDFs and their weights for objects with known 3D geometry~\cite{lensch2001image, lensch2003image, zhou2016sparse}. Moreover, they typically require excessive optimization time and suffer from low accuracy. Recent methods jointly optimize geometry and basis BRDFs~\cite{practical, chung2024differentiable, bi2024rgs}. These approaches often result in non-interpretable basis BRDFs which are spatially entangled: each scene point is represented by many basis BRDFs with high weights and those basis BRDFs are often non-interpretable.
Thus, their results are impractical to use for downstream tasks such as scene editing. Additionally, they are often limited to using a fixed number of basis BRDFs, regardless of scene complexity.

In this paper, we propose a differentiable inverse rendering method that jointly estimates geometry and reflectance. 
We represent geometry as a set of 2D Gaussians~\cite{huang20242d, dai2024high}, each with shape parameters and basis BRDF weights. The reflectance of each Gaussian is modeled as a spatially varying blend of basis BRDFs. We dynamically adjust the number of basis BRDFs during the analysis-by-synthesis optimization, promoting sparsity in the basis BRDF weights. 
To enhance training stability, we employ a weighted photometric loss that focuses on potentially specular regions. 
We demonstrate that our method obtains not only accurate geometry but also scalable and interpretable basis BRDFs which sparsely represent SVBRDFs as shown in Figure~\ref{fig:teaser}. 

In summary, we make the following contributions:
\begin{itemize}
    \item A differentiable inverse rendering method that jointly estimates 2D Gaussians and basis BRDFs, obtaining interpretable basis BRDFs.
    \item A basis BRDF control method and sparsity regularizer that dynamically adjust the number of basis BRDFs during analysis-by-synthesis optimization, promoting interpretability through sparse blending.
    \item Extensive evaluation compared with existing methods, and demonstration of intuitive scene editing.
\end{itemize}
\vspace{+5mm}

\begin{figure*}[!ht]
	\centering
		\includegraphics[width=1\linewidth]{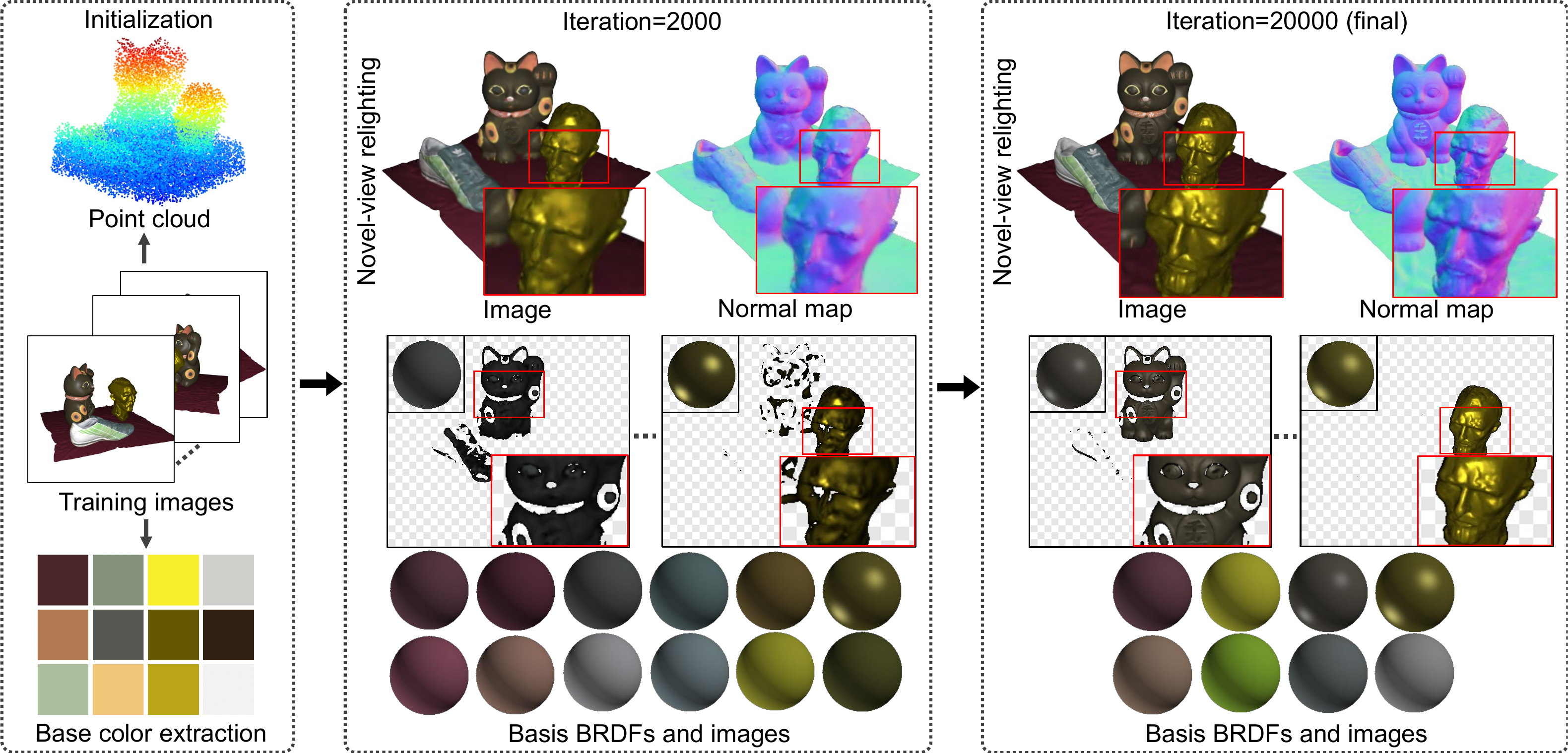}
		\caption{\textbf{The process of our analysis-by-synthesis iterations.} Given a set of multi-view photometric images, we initialize point cloud and extract base color for basis BRDFs. We jointly optimize 2D Gaussians and basis BRDFs by comparing the differentiably-rendered images and the input images. Our method enables obtaining interpretable basis BRDFs with spatially-separated basis-BRDF weights and the number of basis BRDFs adapts to the scene.}
  		\label{fig:pipeline}
\end{figure*}

\section{Related Work}
\label{sec:related}

\paragraph{Inverse Rendering}
Learning-based inverse rendering methods train neural networks such as CNNs~\cite{indoorir, sengupta2019neural, sang2020single, yu2019inverserendernet, li2018learning, wei2020object}, transformers~\cite{zhu2022irisformer}, and diffusion models~\cite{chen2024intrinsicanything, poirier2024diffusion, litman2024materialfusion} on datasets, enabling efficient inference by exploiting prior distributions in the training data. They often suffer from out-of-distribution inputs, resulting in physically inaccurate results.

The analysis-by-synthesis framework enables inverse rendering by inverting the forward rendering process. Recent advancements have been particularly sparked by differentiable rendering techniques using volumetric rendering~\cite{nerd, nerv, nerfactor, psnerf, tensoir, boss2021neural, boss2022samurai, engelhardt2024shinobi, physg, invrender, nero, iron, wu2023nefii, wang2024inverse, fei2024vminer, brument2024rnb, zeng2023nrhints} and rasterization~\cite{gao2023relightable, liang2024gs, jiang2024gaussianshader, shi2023gir, zhu2024gs, bi2024rgs}. Among analysis-by-synthesis approaches, recent rasterization-based methods using isotropic points~\cite{chung2024differentiable}, 2D Gaussians~\cite{zhao2024surfel}, and 3D Gaussians~\cite{gao2023relightable, liang2024gs, jiang2024gaussianshader, shi2023gir, zhu2024gs, bi2024rgs} exhibit both efficiency and accuracy. However, they suffer when insufficient light-view angular samples are provided.
Incorporating basis BRDFs into the modern analysis-by-synthesis framework has shown promising performance, however the reconstructed basis BRDFs are often non-interpretable, limiting their usefulness in downstream tasks such as scene editing~\cite{chung2024differentiable, bi2024rgs}. In contrast, our analysis-by-synthesis method obtains 2D Gaussians with accurate geometry and interpretable basis BRDFs.

\paragraph{Basis BRDF}
Basis BRDFs have often been used as a representation of SVBRDFs, decomposing the SVBRDF into a blend of basis BRDFs. This approach is based on the spatial coherence of reflectance, assuming that the BRDFs of many pixels are similar in real-world scenes~\cite{matusik2003data, matusik2003efficient}. Using basis BRDFs helps reduce the ill-posedness of inverse rendering because it enables gathering many light-angular observations for each basis BRDF reconstruction. Existing methods jointly optimize basis BRDFs and spatially varying weights for planar~\cite{ren2011pocket, trees} and non-planar objects~\cite{alldrin2008photometric, shape, practical, hui2016shape}.
Recently, two methods~\cite{chung2024differentiable, bi2024rgs} demonstrated the effectiveness of basis BRDFs in modern differentiable analysis-by-synthesis frameworks. While they showed promising results, the estimated basis BRDFs are non-interpretable {as shown in Figure~\ref{fig:basis_comparison}}, which limits their utility in downstream tasks such as scene editing. Moreover, generalizing their methods to varying degrees of scene complexity is challenging because the number of basis BRDFs is fixed.
To overcome these problems, we are inspired by the decade-ago work of Zhou et al.~\cite{zhou2016sparse}, which proposed adjusting the number of basis BRDFs during optimization. Their method assumes known geometry, does not exploit a differentiable rendering pipeline, and suffers from low reconstruction accuracy and efficiency. We address these challenges by developing a differentiable inverse rendering method that jointly estimates 2D Gaussians and basis BRDFs to obtain interpretable basis BRDFs, where the number of basis BRDFs are automatically controlled.
\section{Method}
\label{sec:method}
Our method consists of two parts.
In Section~\ref{sec:method1}, we focus on optimizing 2D Gaussians and basis BRDFs without concerning the interpretability of basis BRDFs. Next, we extend the method to obtaining interpretable basis BRDFs in Section~\ref{sec:method2}.
Figure~\ref{fig:pipeline} shows the process of our method over analysis-by-synthesis iterations. We obtain not only accurate geometry but also interpretable basis BRDFs whose number adapts to the scene and the basis BRDFs represent SVBRDFs with spatially-separated basis-BRDF weights.

\subsection{Gaussian Inverse Rendering with Basis BRDFs}
\label{sec:method1}
We introduce our method that jointly estimates geometry and a fixed number of basis BRDFs in a differentiable analysis-by-synthesis manner from multi-view flash-photography images.

\paragraph{Geometry}
We use 2D Gaussians as geometric primitives~\cite{huang20242d}, where each Gaussian $g \in G$ is an elliptical disk, and $G$ is the set of all Gaussians. Each Gaussian $g$ is parameterized by its center location $\mathbf{p} \in \mathbb{R}^{3 \times 1}$, principal vectors $\mathbf{t}\in \mathbb{R}^{3 \times 2}$, and scaling factors $\mathbf{s} \in \mathbb{R}^{2\times1}$. The Gaussian surface normal $\mathbf{n}$ is computed as the cross product of the principal vectors.

\paragraph{Basis BRDFs}
We use a basis BRDF representation, where we define $N$ basis BRDFs. We model the $i$-th basis BRDF $f_{i}$ with base color $\mathbf{b}_{i}$, roughness $\sigma_{i}$, and metallic parameter $m_{i}$ using the simplified Disney BRDF model~\cite{yao2022neilf}:
\begin{align}\label{eq:brdf}
f_{i}(\mathbf{i}, \mathbf{o}) = & \frac{1 - m_{i}}{\pi}  \mathbf{b}_{i}  \nonumber \\
&+ \frac{D(\mathbf{h}; \sigma_{i})  F(\mathbf{o}, \mathbf{h}; \mathbf{b}_{i}, m_{i})  G(\mathbf{i}, \mathbf{o}, \mathbf{n}; \sigma_{i})}{4 (\mathbf{n} \cdot \mathbf{i})(\mathbf{n} \cdot \mathbf{o})},
\end{align}
where $\mathbf{i}$ and $\mathbf{o}$ are the incident and outgoing directions, and $\mathbf{h} = (\mathbf{i} + \mathbf{o}) / 2$ is the half-way vector. $D$ is the normal distribution function, $F$ is the Fresnel term, and $G$ is the geometric attenuation. Details of these terms are provided in the Supplemental Document.

We assign blending weights $w_{i}(g)$ for every $i$-th basis BRDF per each Gaussian $g$, and represent the BRDF $f$ of the Gaussian $g$ as:
\begin{equation}\label{eq:basis_brdf}
f(\mathbf{i}, \mathbf{o}; g) = \sum_{i=1}^{N} w_{i}(g)  f_{i}(\mathbf{i}, \mathbf{o}).
\end{equation}

\paragraph{Scene Representations}
In summary, our scene representation consists of per-Gaussian parameters $G$ and basis BRDF parameters $R$:
\begin{align}
g &= \underbrace{\{ \mathbf{p}, \mathbf{t}, \mathbf{s} \}}_{\text{geometric}} \cup \underbrace{\{\{ w_{i}(g)\}_{i=1}^N, \alpha \}}_{\text{photometric}}, \text{where } g \in G, \nonumber \\
R &=  \{ m_{i}, \mathbf{b}_{i}, \sigma_{i} \}_{i=1}^N.
\end{align}
$\alpha$ is the Gaussian opacity. To initialize the basis BRDF parameters $R$, we perform k-means clustering with a fixed number $N$ on the input multi-view flash photography images $I'$. The base colors $\{ \mathbf{b}_{i} \}_{i=1}^N$ are initialized as the mean values of the clusters. The roughness $\{ \sigma_{i} \}_{i=1}^N$ and metallic parameters $\{ m_{i} \}_{i=1}^N$ are set to initial values: $\sigma_{i} = 0.5$ and $m_{i} = 0.0$. We initialize the weights $w_{i}(g)$ uniformly and set $\alpha$ following Huang et al.~\cite{huang20242d}.

\paragraph{Differentiable Rendering}
We render an image under the flash light, modeled as a point light source, using differentiable rasterization. Specifically, we compute the radiance $L$ of each Gaussian $g$ as:
\begin{equation}
\label{eq:simplerendering}
L(\mathbf{i}, \mathbf{o}; g) = (\mathbf{n} \cdot \mathbf{i})  f(\mathbf{i}, \mathbf{o}; g)  E(g),
\end{equation}
where $E(g)$ is the {incident light intensity} on Gaussian $g$.

After computing the radiance, we rasterize the Gaussians onto the image plane, sort them by depth, and accumulate radiance values using alpha blending~\cite{huang20242d} to produce the final pixel value of the rendered image $I$.
We define this rendering as a function $\mathrm{render}( \cdot )$: 
\begin{align}\label{eq:rendering}
I(u) &= \mathrm{render}\left( \{ L_{i} \}_{i=1}^M \right) \nonumber \\
&= \sum_{i=1}^M L_{i}  \alpha_{i}  \mathcal{G}_{i}(\mathbf{r}(u)) \prod_{j=1}^{i-1} \left( 1 - \alpha_{j}  \mathcal{G}_{j}(\mathbf{r}(u)) \right),
\end{align}
where $u\in U$ is a pixel, $M$ is the number of Gaussians projected onto pixel $u$.
$\{ L_{i} \}_{i=1}^M$ and $\{ \alpha_{i} \}_{i=1}^M$ are the radiance and opacity values of the depth-sorted $i$-th Gaussian, respectively. 
The vector $\mathbf{r}(u)$ is the ray coming from corresponding pixel $u$, and $\mathcal{G}_{i}(\mathbf{r}(u))$ is the Gaussian-filtered distance between the center of $i$-th Gaussian and the intersection point between the ray $\mathbf{r}(u)$ and the $i$-th Gaussian. Details are provided in the Supplemental Document.

\paragraph{Optimization}
We optimize the Gaussian parameters $G$ and the basis BRDF parameters $R$ by minimizing the loss function $\mathcal{L}$ in an analysis-by-synthesis manner:
\begin{equation}\label{eq:loss_function1}
\mathcal{L} = \mathcal{L}_{\text{render}} + \lambda_{\text{geom}}  \mathcal{L}_{\text{geom}} + \lambda_{\text{mask}}  \mathcal{L}_{\text{mask}},
\end{equation}
where $\mathcal{L}_{\text{render}}$ penalizes the difference between the rendered image $I$ and the input image $I'$ across all input views. $\mathcal{L}_{\text{geom}}$ is a geometric regularization term for depth distortion and normal consistency~\cite{huang20242d}, and $\mathcal{L}_{\text{mask}}$ is the cross-entropy loss between the rendered mask and the ground-truth mask. $\lambda_{\text{geom}}$ and $\lambda_{\text{mask}}$ are the balancing weights. Details are provided in the Supplemental Document.

\paragraph{Specular-weighted Rendering Loss}
Multi-view flash photography typically generates fewer observations of specular-dominant pixels compared to diffuse-dominant pixels. To address this imbalance, we weight specular observations by using the potentially-specular weight map:
\begin{equation}\label{eq:weighted_photometric_loss}
H(u) =  1 + \lambda_{\theta_{h}}  (\cos \theta_{h}(u))^{k},
\end{equation}
where $\theta_{h}$ is the rendered half-way angle map as $\theta_{h}(u) = \mathrm{render}\left( \{ \arccos(\mathbf{n}_{i} \cdot \mathbf{h}_{i}) \}_{i=1}^M \right)$. $\mathbf{n}_{i}$ and $\mathbf{h}_{i}$ are the normal and half-way vector of Gaussian $i$, respectively. 
{We set the scalar $\lambda_{\theta_{h}}=5$ and $k=10$ to strongly weight specular regions.

We use the potentially-specular weight map $H$ for calculating the final rendering loss:
\begin{equation}\label{eq:weighted_photometric_loss}
\mathcal{L}_{\text{render}} = \frac{1}{|U|} \sum_{u \in U} H(u) \left( (1 - \lambda_{s})  \mathcal{L}_{1}(u) + \lambda_{s}  \mathcal{L}_{\text{SSIM}}(u) \right),
\end{equation}
where $\mathcal{L}_{1}(u) = \| I(u) - I'(u) \|_{1}$, $\mathcal{L}_{\text{SSIM}}(u)$ is the SSIM loss between $I(u)$ and $I'(u)$, and $\lambda_{s}$ is a balancing weight.

\begin{figure}[t]
\centering
\includegraphics[width=\linewidth]{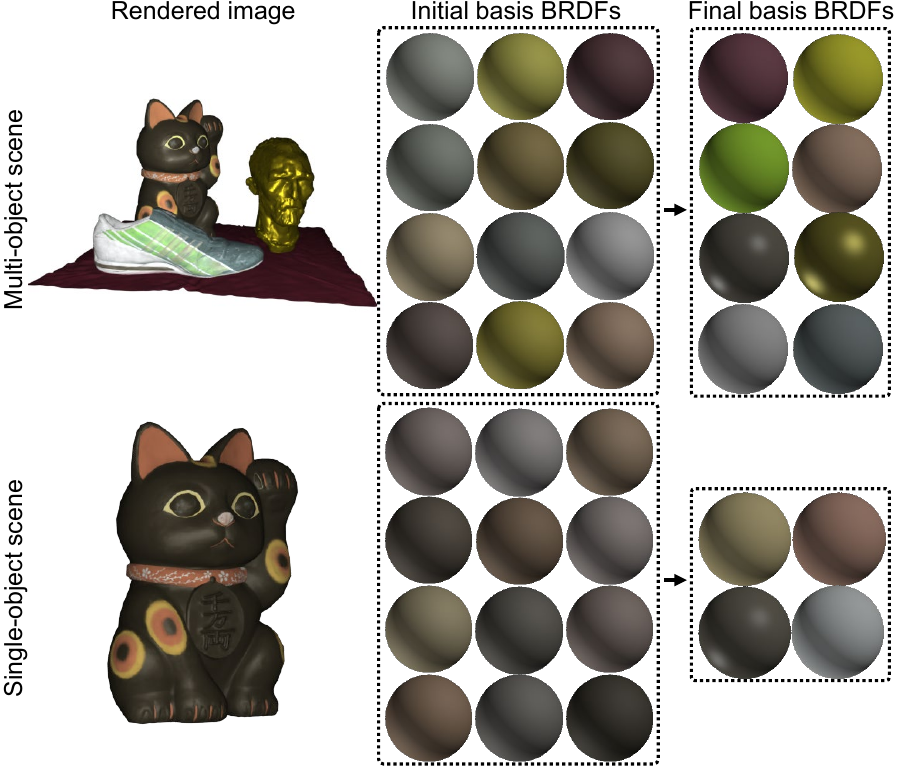}
\caption{\textbf{Basis BRDFs with varying scene complexity.} We adjust the number of basis BRDFs during analysis-by-synthesis iterations to adapt to the scene complexity. We initialize the same number of basis BRDFs for both scenes in this example. }
\vspace{-3mm}
\label{fig:scene_complexity}
\end{figure}

\begin{figure*}[t]
    \centering
    \includegraphics[width=\linewidth]{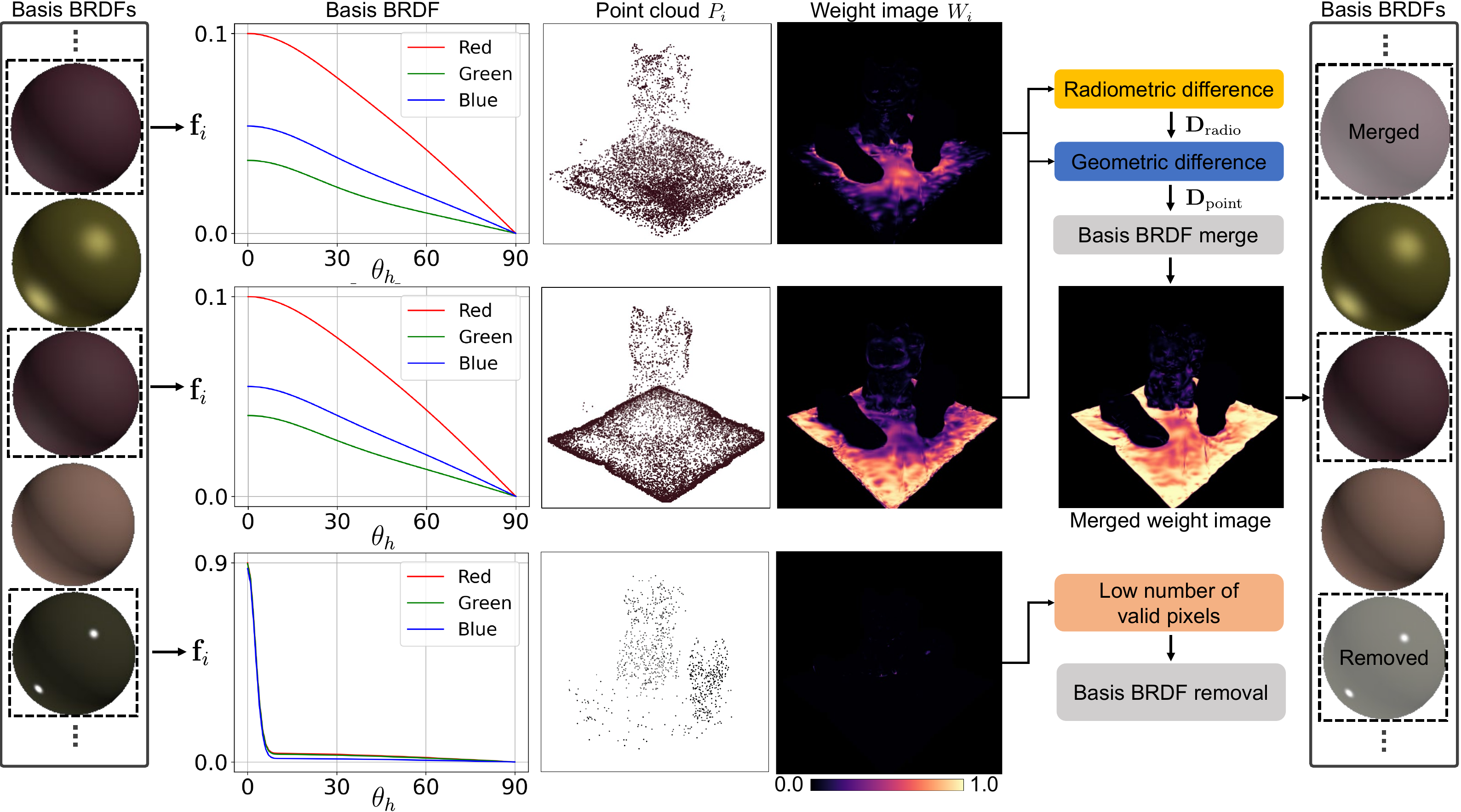}
    \caption{\textbf{Basis BRDF control.} During the analysis-by-synthesis optimization, we compute the values of each basis BRDF for sampled half-way angles $\theta_{h}$ from which radiometric {difference} is obtained. We compute the geometric difference between point clouds of basis BRDFs.
    If two basis BRDFs are radiometrically and geometrically similar, we merge them.
    If the rendered weight map $W_i$ has few valid pixels, we remove the basis BRDF.
}
\vspace{-2mm}
\label{fig:basis_control}
\end{figure*}

\subsection{Interpretable Basis BRDFs}
\label{sec:method2}

While our method in Section~\ref{sec:method1} offers accurate inverse rendering, the reconstructed basis BRDFs lacks of interpretability. {Here, we extend the method to obtain interpretable basis BRDFs whose number adapts to the scene as shown in Figure~\ref{fig:scene_complexity}.}
We develop basis BRDF control method of merge and removal in addition to impose sparsity on basis BRDF weights.

\paragraph{Sparsity of Basis BRDF Weights}
For obtaining interpretable basis BRDF, each scene point needs to be represented with only few basis BRDFs with high weights.
This enables spatially separating SVBRDFs with sparse basis BRDFs.
We impose sparsity on the basis BRDF weights $\{ w_{i}(g) \}_{i=1}^N$ for each Gaussian $g$, making each Gaussian represented by only a few basis BRDFs.

First, we apply a softmax function with a low temperature $T=0.0125$ to the weights, promoting sparse per-Gaussian weights:
\begin{equation}
w_{i}(g) \leftarrow \frac{\exp\left( w_{i}(g) / T \right)}{ \sum_{i'=1}^N \exp\left( w_{i'}(g) / T \right)}.
\end{equation}
Second, we apply an entropy-based sparsity regularizer on the Gaussian weight $w_{i}$ and rendered weight image $W_{i}$ for each $i$-th basis BRDF:
\begin{align}
    \mathcal{L}_{\text{sparse}} = &- \frac{1}{|G|} \sum_{g \in G} \sum_{i=1}^{N} w_{i}(g) \log w_{i}(g) \nonumber \\
    & -\frac{1}{|U|} \sum_{u \in U} \sum_{i=1}^{N} W_{i}(u) \log W_{i}(u),\\
    \label{eq:weight}
    W_{i} =& \mathrm{render}\left( \{ w_{i}(g) \}_{g \in G} \right).
\end{align}

\begin{figure*}[t]
\centering
\includegraphics[width=\linewidth]{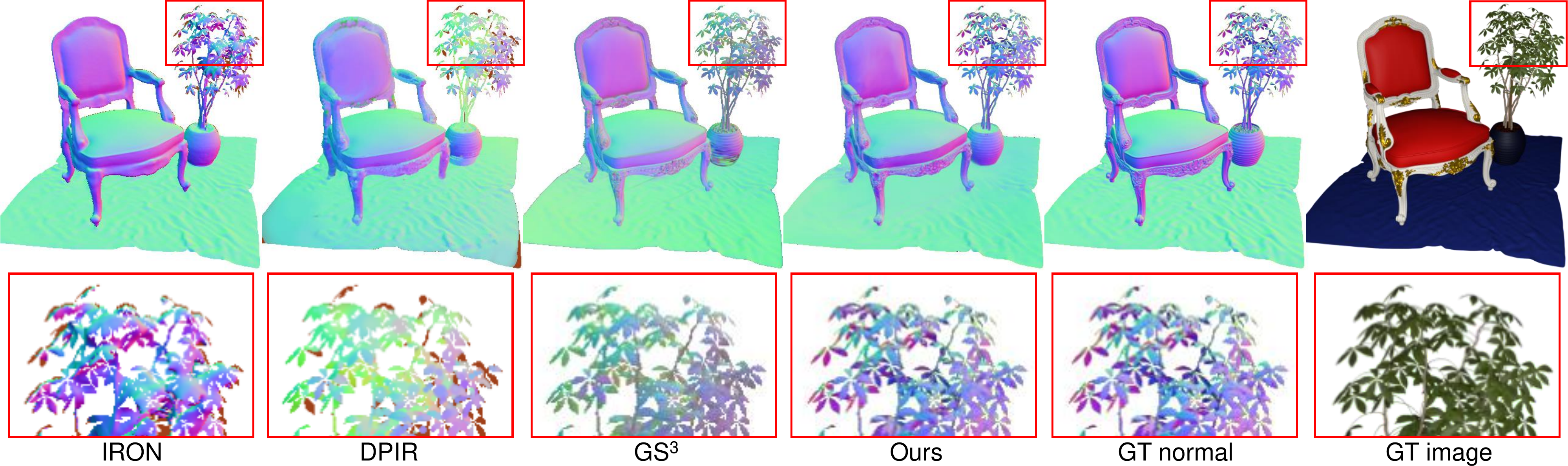}
\caption{\textbf{Normal reconstruction.} Our method successfully recovers the detailed surface normal, outperforming other state-of-the-art inverse rendering methods: IRON~\cite{iron}, DPIR~\cite{chung2024differentiable}, GS\textsuperscript{3}~\cite{bi2024rgs}.}
\vspace{-3mm}
\label{fig:normal_comparison}
\end{figure*}
\paragraph{Basis BRDF Merge}
Basis BRDFs often become similar during optimization, making the resulting basis BRDFs non-intuitive with duplicates. 
We address this by merging similar basis BRDFs during analysis-by-synthesis optimization, as shown in Figure~\ref{fig:basis_control}. We compute the values of each basis BRDF $f_{i}$ at uniformly-sampled halfway angles $\theta_{h}$ for the RGB channels, resulting in a matrix $\mathbf{f}_{i} \in \mathbb{R}^{C \times 3}$. $C$ is the number of halfway angles. We then compute the radiometric {difference} between pairs of basis BRDFs, represented as a matrix $\mathbf{D}_{\text{radio}} \in \mathbb{R}^{N \times N}$, which is defined as:
\begin{equation}\label{eq:similarity_matrix}
\mathbf{D}_{\text{radio}}(i, j) = \frac{1}{C} \left\| \mathbf{f}_{i} - \mathbf{f}_{j} \right\|_{2}.
\end{equation}
We also compute the geometric difference between the every pair of basis BRDFs, represented as a matrix $\mathbf{D}_{\text{point}} \in \mathbb{R}^{N \times N}$ using the Chamfer distance between point clouds represented by each basis BRDF:
\begin{align}
\mathbf{D}_{\text{point}}(i, j)  = & \frac{1}{|P_{i}|} \sum_{\mathbf{p} \in P_{i}} \min_{\mathbf{q} \in P_{j}} \| \mathbf{p} - \mathbf{q} \|_{2}^{2} \nonumber \\
&+ \frac{1}{|P_{j}|} \sum_{\mathbf{q} \in P_{j}} \min_{\mathbf{p} \in P_{i}} \| \mathbf{q} - \mathbf{p} \|_{2}^{2},
\end{align}
where $P_{i}$ and $P_{j}$ are the point clouds for the $i$-th and $j$-th basis BRDFs, consisting of the center locations of Gaussians where the $i$-th or $j$-th basis BRDF has the highest weight.

If the radiometric {difference} $\mathbf{D}_{\text{radio}}(i, j)$ is below a threshold $\tau_{\text{merge}}$ and $\mathbf{D}_{\text{point}}(i, j)$ is minimal among all $\forall j$, we merge the two basis BRDFs by deleting the basis BRDF with fewer associated Gaussians and reassigning the per-Gaussian basis BRDF weights as
\begin{equation}
w_{j}(g) \leftarrow w_{i}(g) + w_{j}(g), \quad \forall g\in G.
\end{equation}

\paragraph{Basis BRDF Removal}
To obtain meaningful basis BRDFs only, we remove $i$-th basis BRDF if it contributes to only a few pixels during optimization as shown in Figure~\ref{fig:basis_control}. We evaluate this using the rendered weight image $W_{i}$ from Equation~\eqref{eq:weight} as:
\begin{equation}\label{eq:delete}
\frac{ |\{ u \in U  |  W_{i}(u) > \tau_{\text{removal-weight}} \}| }{ |U| } < \tau_{\text{removal-number}},
\end{equation}
where $\tau_{\text{removal-weight}}=0.1$ is the threshold determining whether the $i$-th basis BRDF significantly contributes to that pixel $u$, and $\tau_{\text{removal-number}}=0.005$ is the threshold for the normalized number of such pixels.

\paragraph{Scheduling}
For a warm start, we first run our inverse rendering described in Section~\ref{sec:method1}. After a predefined number of iterations, 6000 in our experiments, we perform basis BRDF merge and removal at predefined intervals 500 during iterations to refine the basis BRDFs.
We incorporate the sparsity loss $\mathcal{L}_{\text{sparse}}$ and optimize the scene parameters by minimizing:
\begin{equation}\label{loss_function}
\mathcal{L} = \mathcal{L}_{\text{render}} + \lambda_{\text{geom}} \mathcal{L}_{\text{geom}} + \lambda_{\text{mask}} \mathcal{L}_{\text{mask}} + \lambda_{\text{sparse}} \mathcal{L}_{\text{sparse}},
\end{equation}
where $\lambda_{\text{sparse}}$ is the balancing weight. {Refer to the Supplemental Document for the details on the optimization.}
\section{Results}
\label{sec:Experiment}
\paragraph{Datasets}
We evaluate our approach on both synthetic and real-world photometric dataset. For the synthetic dataset, we render four complex scenes containing multiple objects, following the co-located flash photography. Each scene includes multiple objects. The dataset consists of 200 training images and 100 test images per scene.
For the real-world photometric dataset, we capture one multi-object scene using a mobile phone with flash at {200/40} training/testing images. We conduct geometric and radiometric calibrations of the light and the camera.

\subsection{Validation and Comparison}
\paragraph{Geometry}
We assess our method in comparison with state-of-the-art analysis-by-synthesis inverse rendering methods: IRON~\cite{iron}, DPIR~\cite{chung2024differentiable} and GS\textsuperscript{3}~\cite{bi2024rgs}. Figure~\ref{fig:normal_comparison} and Table~\ref{tab:normal_table} show the results for geometry reconstruction, demonstrating the highest accuracy in normal reconstruction {with mean angular error (MAE).}
IRON and DPIR employ neural signed distance functions to recover smooth surface normal, not only resulting in loss of details but also requiring extended training times for optimization as shown in Table~\ref{tab:normal_table}.
GS\textsuperscript{3}, which is based on 3D Gaussian splatting with anisotropic spherical Gaussians, exhibits limited accuracy in shape reconstruction. Our method accurately models surface geometry and reflectance using 2D Gaussians and interpretable basis BRDFs, achieving the highest accuracy of normal reconstruction, especially for thin and convex objects. Moreover, we outperform the others in terms of training speed.

\begin{figure*}[t]
\centering
\includegraphics[width=\linewidth]{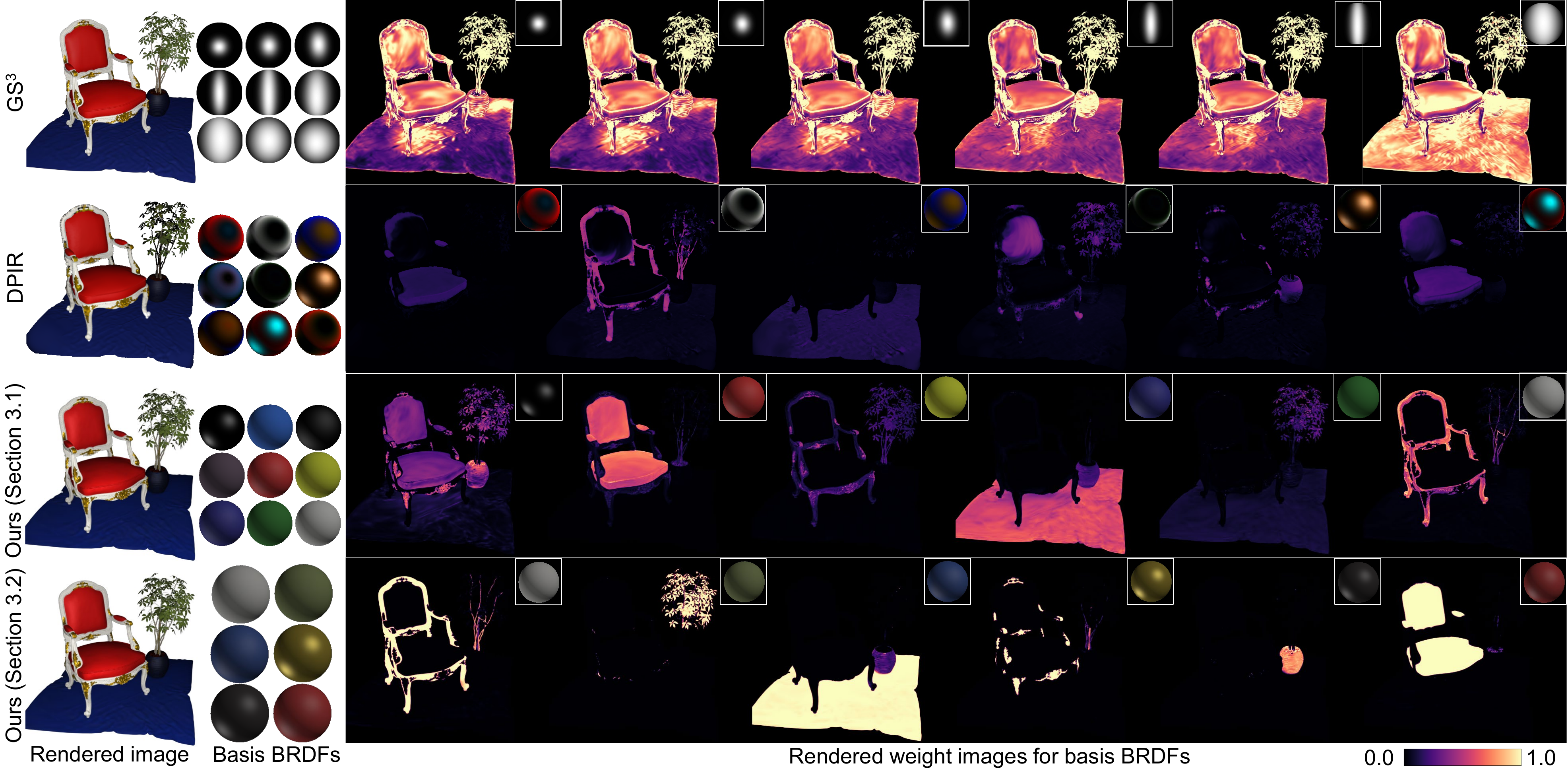}
\caption{\textbf{Basis BRDFs and weight maps.} Our inverse-rendering methods (Section~\ref{sec:method1} and Section~\ref{sec:method2}) obtain basis BRDFs that are explainable with spatially separated weight maps and intuitive basis BRDFs, compared to GS\textsuperscript{3}~\cite{bi2024rgs} and DPIR~\cite{chung2024differentiable}. 
}
\vspace{-2mm}
\label{fig:basis_comparison}
\end{figure*}

\begin{figure*}[t]
\centering
\includegraphics[width=\linewidth]{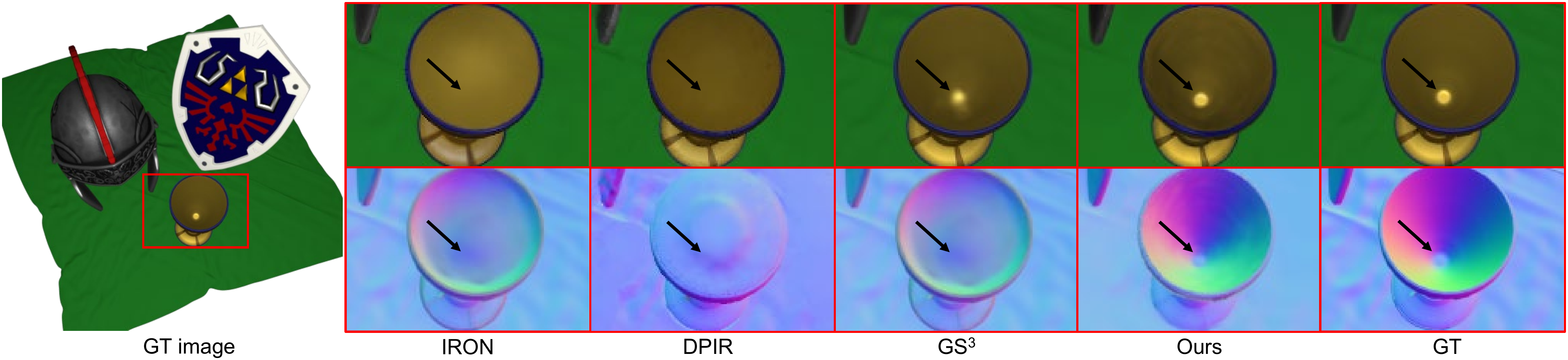}
\caption{\textbf{Novel-view relighting.} Our method enables high quality novel-view relighting with the accurate surface normal. We achieve physically-valid relighting with accurate normal and basis BRDFs.}
\vspace{-3mm}
\label{fig:relighting_comparison}
\end{figure*}

\begin{table}[t]
    \centering
    \resizebox{0.7\columnwidth}{!}{
    \begin{tabular}{c|ccc|c}
        \toprule[1pt]
             & IRON\ &  DPIR & GS\textsuperscript{3} & Ours \\ \hline
        MAE $\downarrow$      &   15.71     &  17.04   &   15.28  &\textbf{9.81} \\
        Train $\downarrow$      &  12h   &  2h    &   1h   &  \textbf{0.5h}\\ 
        \bottomrule[1pt]  
        \end{tabular}}
        \caption{\textbf{Normal reconstruction and training time.} Our method outperforms IRON~\cite{iron}, DPIR~\cite{chung2024differentiable}, and GS\textsuperscript{3}~\cite{bi2024rgs} in quantitative numbers evaluated on the synthetic dataset.}
    \vspace{-2mm}
    \label{tab:normal_table}
\end{table}

\paragraph{Basis BRDF}
Figure~\ref{fig:basis_comparison} shows the estimated basis BRDFs and weight maps, comparing modern inverse-rendering methods using basis BRDFs: GS\textsuperscript{3}, DPIR, our method without interpretable basis BRDFs (Section~\ref{sec:method1}), and our method with interpretable basis BRDFs (Section~\ref{sec:method2}). Both DPIR and GS\textsuperscript{3} use basis BRDFs to leverage spatial coherence of specular reflectance. These methods yield non-interpretable basis BRDFs, as evident from the rendered weight maps and basis BRDF renderings. 
Our method introduced in Section~\ref{sec:method1} improves the interpretability of the basis BRDFs by jointly optimizing base colors.
However, the resulting basis BRDFs are still often duplicated and non-intuitively classified. 
Moreover, all the methods have a fixed number of basis BRDFs, limited to modeling different complexity of scenes.
In contrast, our full method (Section~\ref{sec:method2}) obtains interpretable basis BRDFs whose number fits to the target scene, as shown in Figure~\ref{fig:scene_complexity}.

\subsection{Applications}
\paragraph{Novel-view Relighting}
Figure~\ref{fig:relighting_comparison} shows novel-view relighting and estimated normals, compared with IRON, DPIR, GS\textsuperscript{3}.
They suffer from inaccurate estimation of geometry.
Interestingly, despite inaccurate normals, GS\textsuperscript{3} manages to render high-quality relighting results. This is because neural networks compensate for such imperfection by learning residuals, as relighting is the main goal of GS\textsuperscript{3}. In contrast, our method reconstructs accurate geometry and interpretable basis BRDFs that enable not only novel-view relighting but also downstream tasks such as intuitive scene editing.
Figure~\ref{fig:point_relighting} shows high-quality relighting results of a real-world scene under different point light locations.
For more results, refer to the Supplemental Document.

\begin{figure}[t]
\centering
\includegraphics[width=\linewidth]{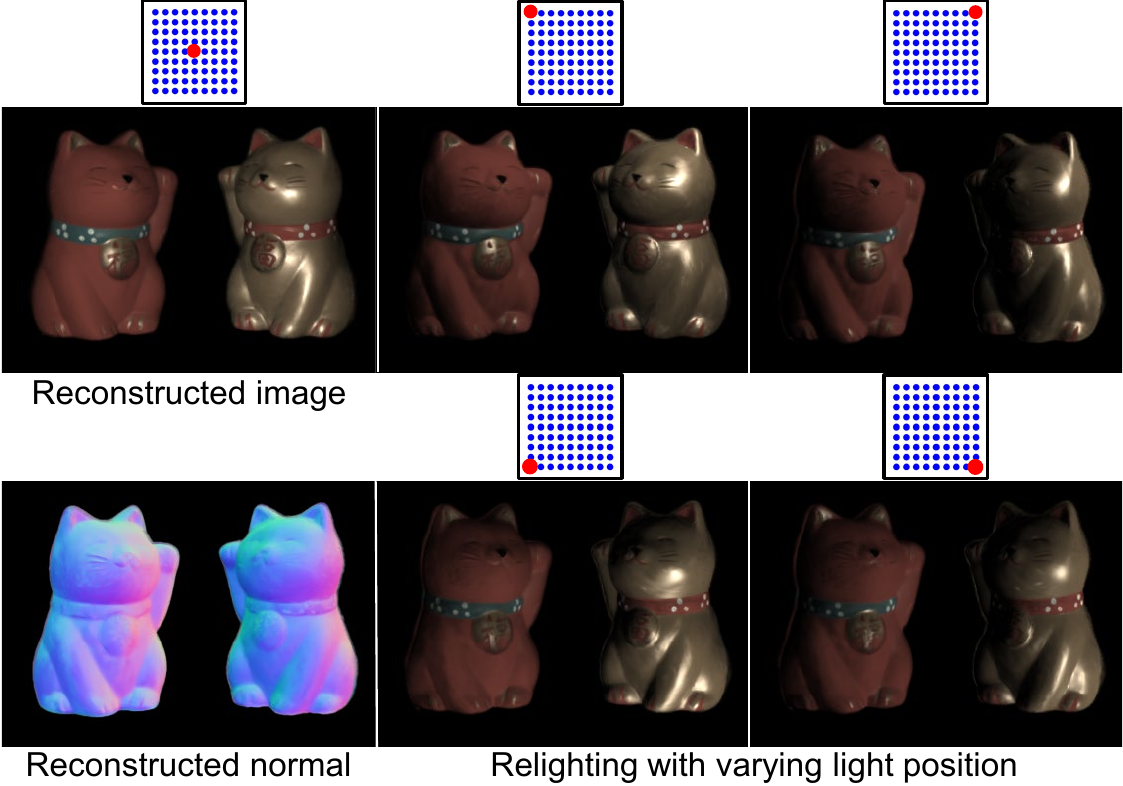}
\caption{\textbf{Relighting with point light sources.} We achieve high-quality relighting with a moving point light source.}
\vspace{-2mm}
\label{fig:point_relighting}
\end{figure}

\paragraph{Scene Editing}

Our inverse rendering method enables selective editing based on basis BRDFs and weight images.
Figure~\ref{fig:scene_editing} shows that our interpretable basis BRDFs facilitate reflectance editing.
We modify the basis BRDF parameters of base color, roughness, and metallic: gold glass to silver glass.
We then obtain a single object from the entire scene by selecting basis BRDFs that are utilized to represent the corresponding object. Next, we delete other basis BRDFs and Gaussians that have the highest weight for other basis BRDFs.
We also filter out Gaussians based on Gaussian positions. 
This allows for extracting object mesh~\cite{huang20242d} and relighting it under an environment map. 

\begin{figure}[t]
\centering
\includegraphics[width=\linewidth]{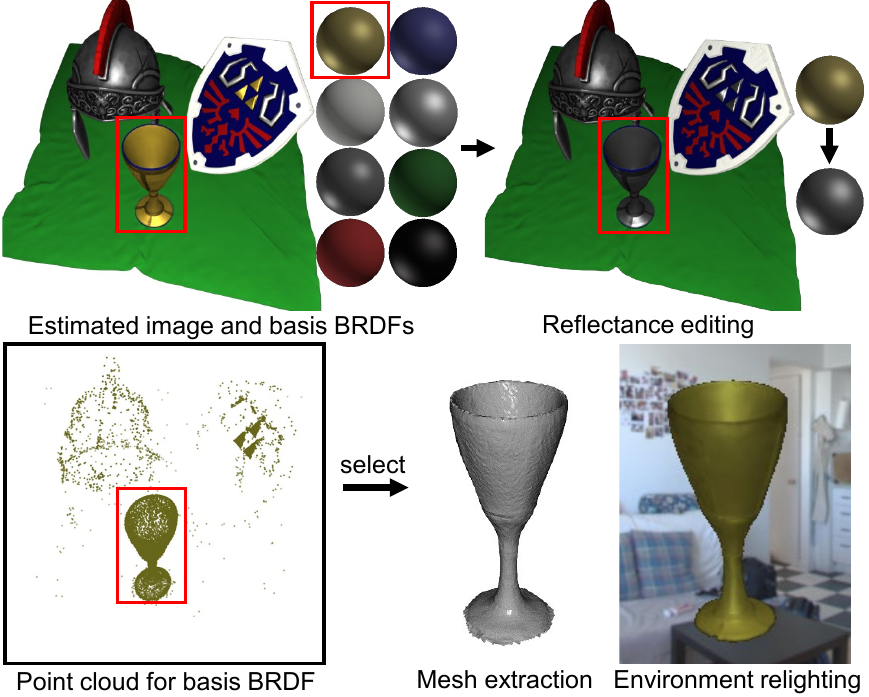}
\caption{\textbf{Intuitive scene editing.} We modify our basis BRDF parameters to edit the reflectance. We also selectively extract mesh and relight it under the environment map.}
\vspace{-3mm}
\label{fig:scene_editing}
\end{figure}
\subsection{Ablation Study}

\begin{figure}[t]
\centering
\includegraphics[width=\linewidth]{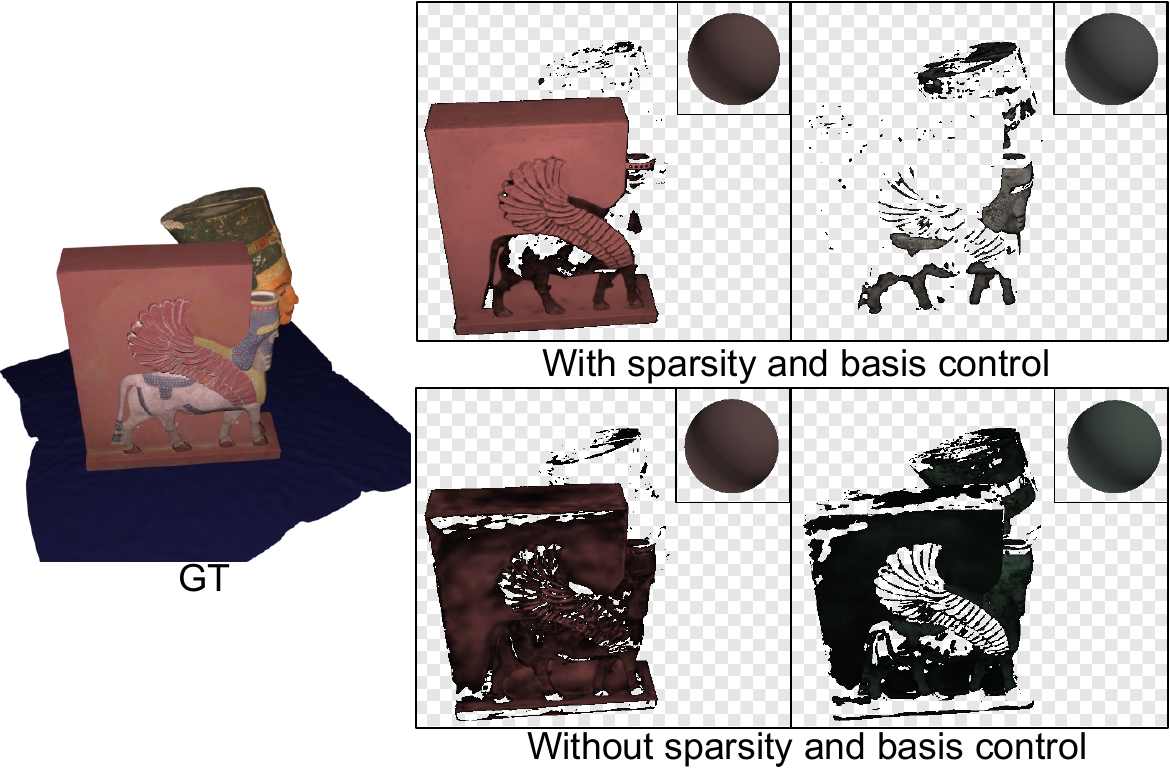}
\caption{\textbf{Impact of sparsity regularizer and basis BRDF control.} Without sparsity loss and basis BRDF control of merge and removal, the resulting basis BRDFs are blended each other as shown in the corresponding images, making them non-interpretable for downstream tasks.}
\vspace{-1mm}
\label{fig:ablation_sparsity}
\end{figure}

\paragraph{Sparsity Regularizer $\mathcal{L}_{\text{sparse}}$}
Figure~\ref{fig:ablation_sparsity} shows that the sparsity regularizer and the basis BRDF control enable obtaining interpretable basis BRDFs.

\paragraph{Specular-weighted Rendering Loss}
We test the impact of the specular-weighted rendering loss that prioritizes potentially-specular areas using the rendered half-way angle map.
Figure~\ref{fig:ablation_specular} shows that specular-weighted rendering loss produces a more stable reconstruction compared to not using the weighting.

\begin{figure}[t]
\centering
\includegraphics[width=\linewidth]{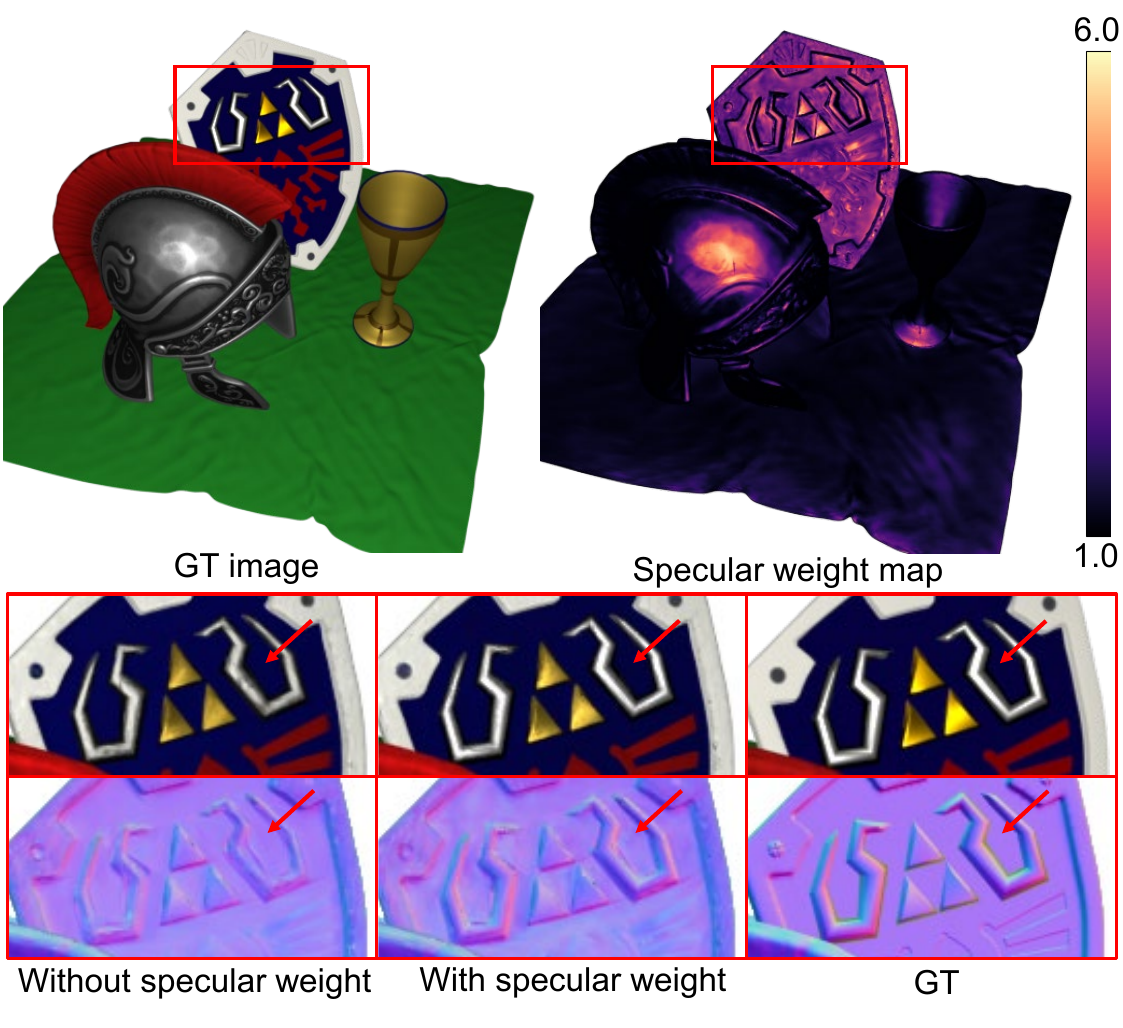}
\caption{\textbf{Impact of specular-weighted loss.} Using specular-weighted rendering loss improves geometry and reflectance reconstruction quality by recovering high frequency features from specularity, which is emphasized by the specular weight map.}
\vspace{-2mm}
\label{fig:ablation_specular}
\end{figure}

\section{Conclusion}
\label{sec:Conclusion}
In this paper, we have introduced a differentiable inverse rendering method that jointly estimates 2D Gaussians and basis BRDFs. Our method obtains interpretable basis BRDFs that can represent SVBRDFs in a spatially-separable manner, enabling accurate geometry and SVBRDF reconstruction.
Also, the number of basis BRDFs automatically adjusts to a target scene. 
We demonstrate the effectiveness of our method compared to state-of-the-art approaches.
Using the interpretable basis BRDFs, our method also facilitates downstream tasks such as intuitive scene editing.

\clearpage
{\small
\bibliographystyle{ieee_fullname}
\bibliography{references}
}}
\end{CJK}
\end{document}


\begin{CJK}{UTF8}{}
\CJKfamily{mj}

\title{Differentiable Inverse Rendering with Interpretable Basis BRDFs

-Supplemental Document-}

\author{
Hoon-Gyu Chung ~ ~ ~
Seokjun Choi ~ ~ ~
Seung-Hwan Baek \\
POSTECH\\
}

\maketitle

In this supplemental document, we provide additional results and details in support of our findings in the main manuscript. 

\tableofcontents

\newpage

\section{Additional Details for the Proposed Method}
\subsection{Simplified Disney BRDF Model}
We use the simplified Disney BRDF model~\cite{yao2022neilf} having parameters: $\mathbf{b}_{i}$, roughness $\sigma_{i}$, and metallic $m_{i}$. The BRDF model has three major terms: distribution function $D$, Fresnel term $F$, and geometry attenuation term $G$ to represent realistic specularity as follows:
\begin{align}\label{eq:brdf}
f_{i}(\mathbf{i}, \mathbf{o}) = & \frac{1 - m_{i}}{\pi}  \mathbf{b}_{i} + \frac{D(\mathbf{h}; \sigma_{i})  F(\mathbf{o}, \mathbf{h}; \mathbf{b}_{i}, m_{i})  G(\mathbf{i}, \mathbf{o}, \mathbf{n}; \sigma_{i})}{4 (\mathbf{n} \cdot \mathbf{i})(\mathbf{n} \cdot \mathbf{o})}.
\end{align}
We compute normal distribution function D using a Spherical Gaussian function:
\begin{equation}
D\left(\mathbf{h};\sigma_{i}\right)=S(\mathbf{h};\textbf{n},\frac{2}{\sigma_{i}\textsuperscript{2}},\frac{1}{\pi \sigma_{i}\textsuperscript{2}})=\frac{1}{\pi \sigma_{i}\textsuperscript{2}} \exp\left(\frac{2}{\sigma_{i}\textsuperscript{2}}(\mathbf{h}\cdot\mathbf{n}-1)\right).
\end{equation}
We use Schlick's approximation for the Fresnel term:
\begin{equation}
F(\mathbf{o}, \mathbf{h}; \mathbf{b}_{i}, m_{i}) = F_{0}+(1-F_{0})\left(1-\mathbf{o}\cdot\mathbf{h}\right)\textsuperscript{5}
\end{equation}
where $F_{0} = 0.04(1-m_{i})+m_{i}\mathbf{b}_{i}$. The geometry attenuation term is computed by multiplication of two GGX functions~\cite{walter2007microfacet}:
\begin{equation}
G(\mathbf{i}, \mathbf{o}, \mathbf{n}; \sigma_{i}) = G_{GGX}(\mathbf{i}\cdot \mathbf{n})G_{GGX}(\mathbf{o}\cdot \mathbf{n}),
\end{equation}
where the GGX function is defined as follows:
\begin{equation}
G_{GGX}(z) = \frac{z}{(1-\alpha)z+\alpha}, ~~ \alpha=\frac{(1+\sigma_{i})\textsuperscript{2}}{8}
\end{equation}

\subsection{2D Gaussian Splatting}
We use 2D Gaussians~\cite{huang20242d} with learnable geometric parameters: $\left\{\mathbf{p},\mathbf{t},\mathbf{s}\right\}$. Each 2D Gaussian is defined on the local tangent frame in world coordinate, which is parameterized as:
\begin{equation}\label{2d_gaussian}
\mathit{P}\!\left(\mathit{a}, \mathit{b}\right)\! = \mathbf{p} + s_{a}\mathbf{t}_{a} \mathit{a} + s_{b}\mathbf{t}_{b} \mathit{b}  = 
\begin{bmatrix}
 \mathbf{t}\mathbf{s}&  \mathbf{p} \\
 \mathbf{0} & \mathbf{1}  \\
\end{bmatrix} 
\left (\mathit{a},\mathit{b},1,1 \right )^{\top},
\end{equation}
where $\mathbf{p}$ is the location of Gaussian center, $\mathbf{t} = \left[\mathbf{t}_a,\mathbf{t}_b,\mathbf{t}_c\right] \in\mathbb{R}^{3\times3}$ denotes the rotation matrix, and $\mathbf{s} = \textrm{diag}\left(s_a,s_b,0\right)\in\mathbb{R}^{3\times3}$ denotes the scaling matrix of 2D Gaussian. $\mathbf{t}_{c}$ denotes the surface normal of 2D Gaussian.
Then, 2D Gaussian-filtered distance value is computed as:
\begin{equation}\label{2d_gaussian_value}
\mathcal{G}(\mathbf{a}) = \exp\left(-\frac{a^2 + b^2}{2}\right),
\end{equation}
where $\textbf{a}=\left(\mathit{a}, \mathit{b}\right)$ denotes a point that is intersected with the ray $\mathbf{r}(u)$ coming from corresponding pixel $u$ and Gaussian. A point $\mathbf{a}$ is defined on the local tangent frame in $ab$ space. Then, we render an image with following equation:
\begin{align}\label{eq:rendering}
I(u) &=  \sum_{i=1}^M L_{i}  \alpha_{i}  \mathcal{G}_{i}(\mathbf{r}(u)) \prod_{j=1}^{i-1} \left( 1 - \alpha_{j}  \mathcal{G}_{j}(\mathbf{r}(u)) \right),
\end{align}
where $u\in U$ is a pixel, $M$ is the number of Gaussians projected onto pixel $u$.
$\{ L_{i} \}_{i=1}^M$ and $\{ \alpha_{i} \}_{i=1}^M$ are the radiance and opacity values of the depth-sorted $i$-th Gaussian, respectively.

\subsection{Loss Functions}
We optimize Gaussian parameters $G$ and basis BRDF parameters $R$ by minimizing the following loss function:
\begin{equation}\label{loss_function}
\mathcal{L} = \mathcal{L}_{\text{render}} + \lambda_{\text{geom}} \mathcal{L}_{\text{geom}} + \lambda_{\text{mask}} \mathcal{L}_{\text{mask}} + \lambda_{\text{sparse}} \mathcal{L}_{\text{sparse}}.
\end{equation}

\begin{itemize}
\item $\mathcal{L}_{\text{render}}$ is the specular-weighted rendering loss consisting of $\mathbf{l}_{1}$ loss and SSIM loss between the reconstructed image $I$ and the observed image $I^{'}$ as follows:
\begin{equation}
\mathcal{L}_{\text{render}} = \frac{1}{|U|} \sum_{u \in U} H(u) \left( (1 - \lambda_{s})  \mathcal{L}_{1}(u) + \lambda_{s}  \mathcal{L}_{\text{SSIM}}(u) \right),
\end{equation}
where $\lambda_{\text{s}}$ is a balancing weight and $H(u)$ is a specular weight map.

    \item To promote stable optimization, we incorporate geometric regularization term $\mathcal{L}_{geom}$ following the approach in 2DGS~\cite{huang20242d}, specifically focusing on depth distortion and normal consistency. Depth distortion loss aims to compact the Gaussian splats by minimizing the depth distance between ray-splat intersections. Normal consistency loss aligns the 2D splats with the object's surface to approximate a smooth surface. The geometric regularization is formulated as:
\begin{equation}\label{sparsity}
\mathcal{L}_\text{geom}\!=\!\sum_{i,j}t_{i}t_{j}\left|z_{j}-z_{i}\right| + \sum_{i=1}t_{i}(1-\mathbf{n}_{i}^{T}\tilde{\mathbf{n}_{i}})
\end{equation}
where $i$ indexes the splats intersected splats along the ray, $t_{i}=\alpha_i \mathcal{G}_i(\mathbf{r}(u)) \prod_{j=1}^{i-1} \left( 1 - \alpha_j \mathcal{G}_j(\mathbf{r}(u)) \right)$, $z_{i}$ is the depth of the splat, $\mathbf{n}_{i}$ is the normal of the rendered normal image, and $\tilde{\mathbf{n}_{i}}$ is the normal computed from the rendered depth image. 
\item $\mathcal{L}_{\text{mask}}$ is the cross entropy loss between rendered mask $M$ and ground-truth mask $M^{'}$. 
\item To obtain spatially separated interpretable basis BRDFs, we apply sparsity regularizer $\mathcal{L}_{\text{sparse}}$ with Gaussian weights and weight images. 
\end{itemize}

\subsection{Point Initialization}
For the synthetic dataset, we utilize masked-based point sampling for the initial point cloud. We sample 3D points uniformly and project these points on the image plane to estimate points that fall within masked region~\cite{point_radiance}. We employ COLMAP output as a point cloud initialization for real-world photometric images. This point cloud initialization enables stable and accurate reconstruction of 3D geometry. 

\subsection{Basis BRDF Initialization}
For basis BRDF initialization, we perform k-means clustering to obtain the initial base color for each basis. We utilize all input pixel values as input and compute the centers of clusters with the number of initialized basis BRDFs. We set $\mathbf{b}_{i}$ as center of cluster, $\sigma_{i}=0.5$ and $m_{i}=0.0$.
This basis BRDF initialization enables efficient and accurate reconstruction of spatially-varying BRDFs with interpretable basis BRDFs.
\subsection{Adaptive Density Control}
We follow the adaptive density control policy of 2D Gaussian splatting~\cite{huang20242d}. We repeat point upsampling and pruning based on the gradient of Gaussian parameters to achieve high-quality reconstruction. To optimize stable geometry and interpretable basis BRDFs efficiently, we increase the densification interval to 500.

\subsection{Scheduling of Loss Functions}
We optimize Gaussian and basis BRDF parameters simultaneously without sparsity regularizers before 5000 iterations. We apply Gaussian weight sparsity regularization after 5000 iterations and weight image sparsity regularization after 9000 iterations. Weight image regularization constraints the sparsity of basis BRDF weights stronger than Gaussian weight regularization.

\section{Dataset}
\subsection{Synthetic Photometric Dataset}
We test our method on synthetic photometric dataset, following the configuration of mobile flash photography ~\cite{practical}. We rendered 4 complex scenes of multiple objects with Blender using mesh and image texture data from IRON~\cite{iron} and Objaverse~\cite{objaverse}. We rendered 300 view images with co-located point lights and used 200/100 views for training/testing, respectively. We also render ground-truth normal images for normal evaluation.

\subsection{Real-world Photometric Dataset}
We test our method on real-world photometric dataset, which is captured by a mobile phone with flashlight. We utilize Proshot application to capture raw images in dng format. We fix exposure, focal length, ISO and white balance to capture images with consistent settings. We perform checkerboard calibration to obtain camera intrinsic parameters and rectify distorted images. We also capture color checker images to estimate white balance and flash light intensity simultaneously. Chrome ball calibration~\cite{lensch2003image} is also performed to estimate the accurate position of flash light, assuming a point light source.

\section{Additional Ablation Study}
\begin{table*}[h]
\centering
\resizebox{0.45\linewidth}{!}{
    \begin{tabular}{c|cccc}
    \toprule[1pt]
    &\multicolumn{4}{c}{Synthetic photometric dataset} \\ \hline
    Method &PSNR $\uparrow$&SSIM $\uparrow$&LPIPS $\downarrow$   &MAE $\downarrow$    \\
    \hline
    Proposed   & 31.78 & 0.9754 & 0.0234 & 9.81 \\ \hline
    w/o spec    & 30.82 & 0.9748 & 0.0299 & 10.29 \\
    w/o softmax   & 30.19 & 0.9720 & 0.0381  & 10.17  \\ 
    w/o $\mathcal{L}_\text{sparse}$  & 34.13 & 0.983 & 0.0180 & 9.85  \\
    w/o $\mathcal{L}_\text{mask}$  & 32.51 & 0.9790 & 0.0186 & 11.05  \\
    basis 9  & 31.51 & 0.9727 & 0.0257 & 10.21    \\ 
    basis 15   & 32.13 & 0.9744 & 0.0242 & 9.82 \\ 
    \bottomrule[1pt]
    \end{tabular}}
    \caption{\textbf{Quantitative comparison of ablation studies for a synthetic photometric dataset.} We achieve the highest reconstruction quality without sparsity loss, however it produces non-interpretable basis BRDFs and weight maps. Our proposed method enables reconstruction of both high-fidelity and interpretable basis BRDFs simultaneously.}
\label{tab:ablation_table}
\vspace{-3mm}
\end{table*}

\subsection{Specular-weighted Photometric Loss}
We evaluate the importance of specular-weighted photometric loss of our method. Table~\ref{tab:ablation_table} shows that a specular-weight map improves both image and normal reconstruction quality. Especially, the specular region is weighted by potentially-specular weight map, resulting in an accurate reconstruction of specularity.

\subsection{Softmax Function with Temperature}
\begin{figure}[h]
    \centering
    \includegraphics[width=0.95\linewidth]{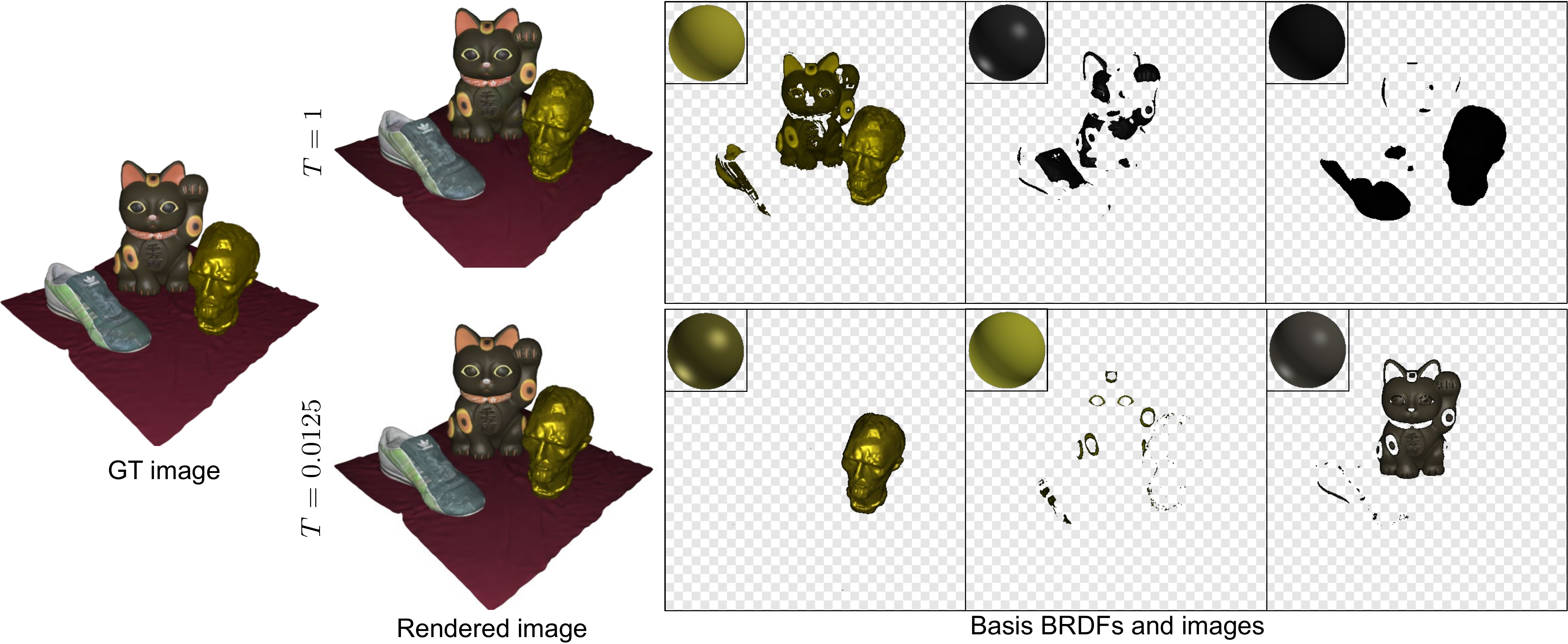}
    \caption{\textbf{Impact of softmax function with low temperature.} Low temperature softmax function results in interpretable basis BRDFs and images. We employ $T=0.0125$ in our experiment.}
\label{fig:ablation_softmax} 
\end{figure}

We evaluate the impact of softmax function with low temperature. Table~\ref{tab:ablation_table} and Figure~\ref{fig:ablation_softmax} show that high temperature $T=1$, results in non-interpretable basis BRDFs and non-intuitive weight images. We employ low temperature softmax function to optimize spatially separated weight maps with interpretable basis BRDFs, enabling intuitive scene editing.

\subsection{Sparsity Regularizer}
Table~\ref{tab:ablation_table} shows that sparsity regularizer downgrades the reconstruction quality while it produces interpretable basis BRDFs. Without sparsity regularizer, optimized basis BRDFs are non-interpretable, and weight images are not spatially separated.

\subsection{Mask Loss}
\begin{figure}[h]
    \centering
    \includegraphics[width=0.95\linewidth]{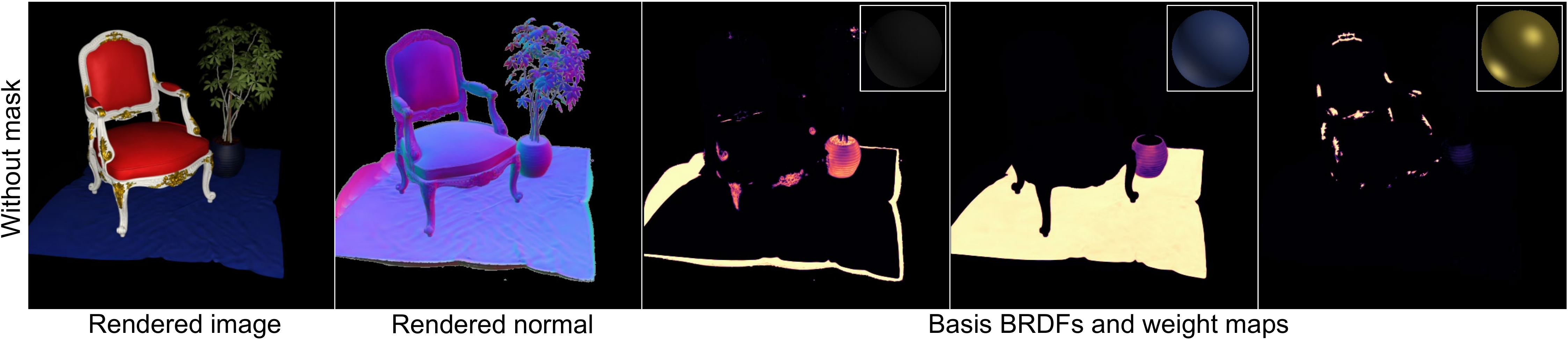}
    \caption{\textbf{Impact of the mask input.} Without mask loss, weight maps are non-intuitive, and the edge is unclear.}
\label{fig:ablation_mask} 
\end{figure}

Table~\ref{tab:ablation_table} shows that we achieve high-quality reconstruction results without mask loss. However, we obtain non-interpretable basis BRDFs and weight maps as shown in Figure~\ref{fig:ablation_mask}. Mask loss regularizes Gaussians to align with the object surface that edge does not include the color of the background. We apply mask loss to reconstruct more interpretable basis BRDFs and accurate surface normal.

\subsection{Number of Initialized Basis BRDFs}
\begin{figure}[h]
    \centering
    \includegraphics[width=0.95\linewidth]{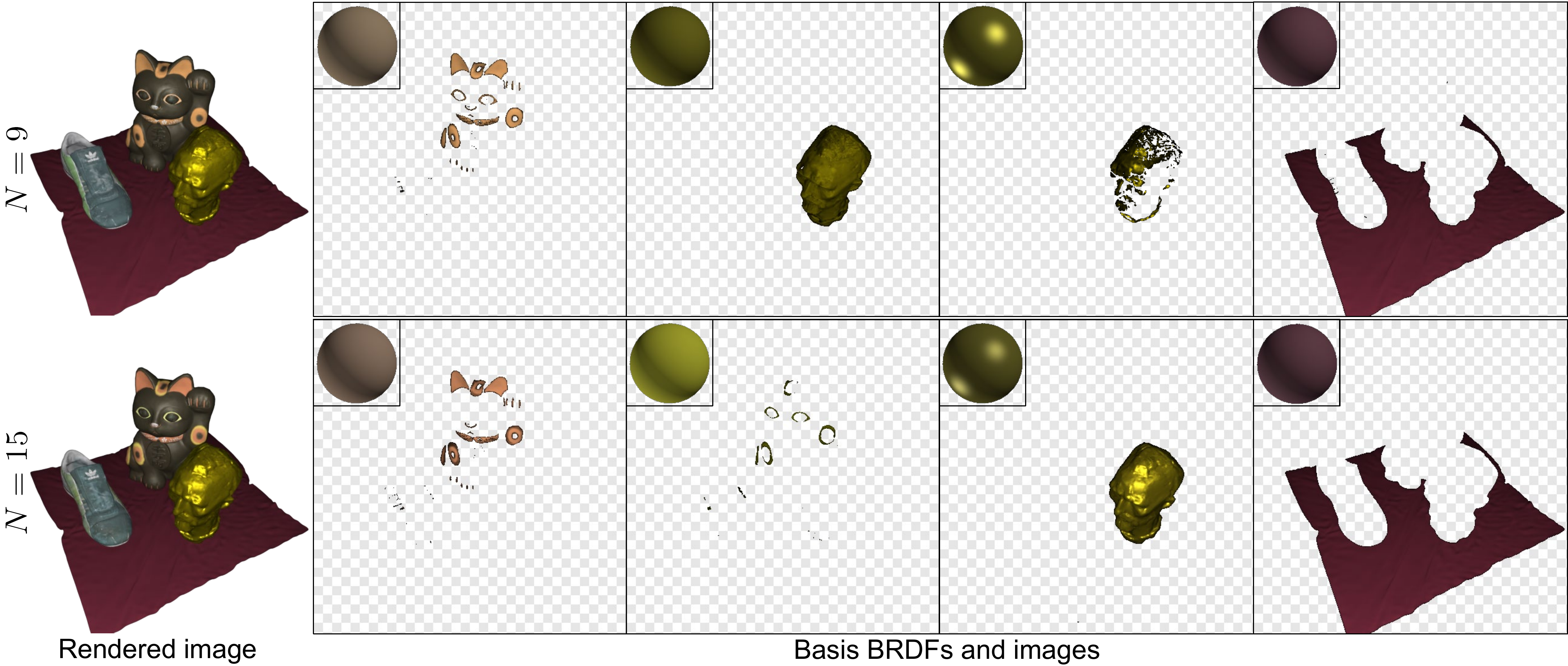}
    \caption{\textbf{Impact of the number of initialized basis BRDFs.} With a small number of initialized basis BRDFs, basis BRDFs are non-interpretable, and weight maps are not spatially separated. We utilize $N=12$ in our experiment.}
\label{fig:ablation_number} 
\end{figure}

We evaluate the impact of the number of initialized basis BRDFs. We test our method on complex scenes with multi-object which need sufficient basis BRDFs. Table~\ref{tab:ablation_table} and Figure~\ref{fig:ablation_number} show that using a small number of basis BRDFs results in low reconstruction quality and non-interpretable basis BRDFs.

\subsection{Threshold for Basis BRDF Merge}
\begin{figure}[h]
    \centering
    \includegraphics[width=0.95\linewidth]{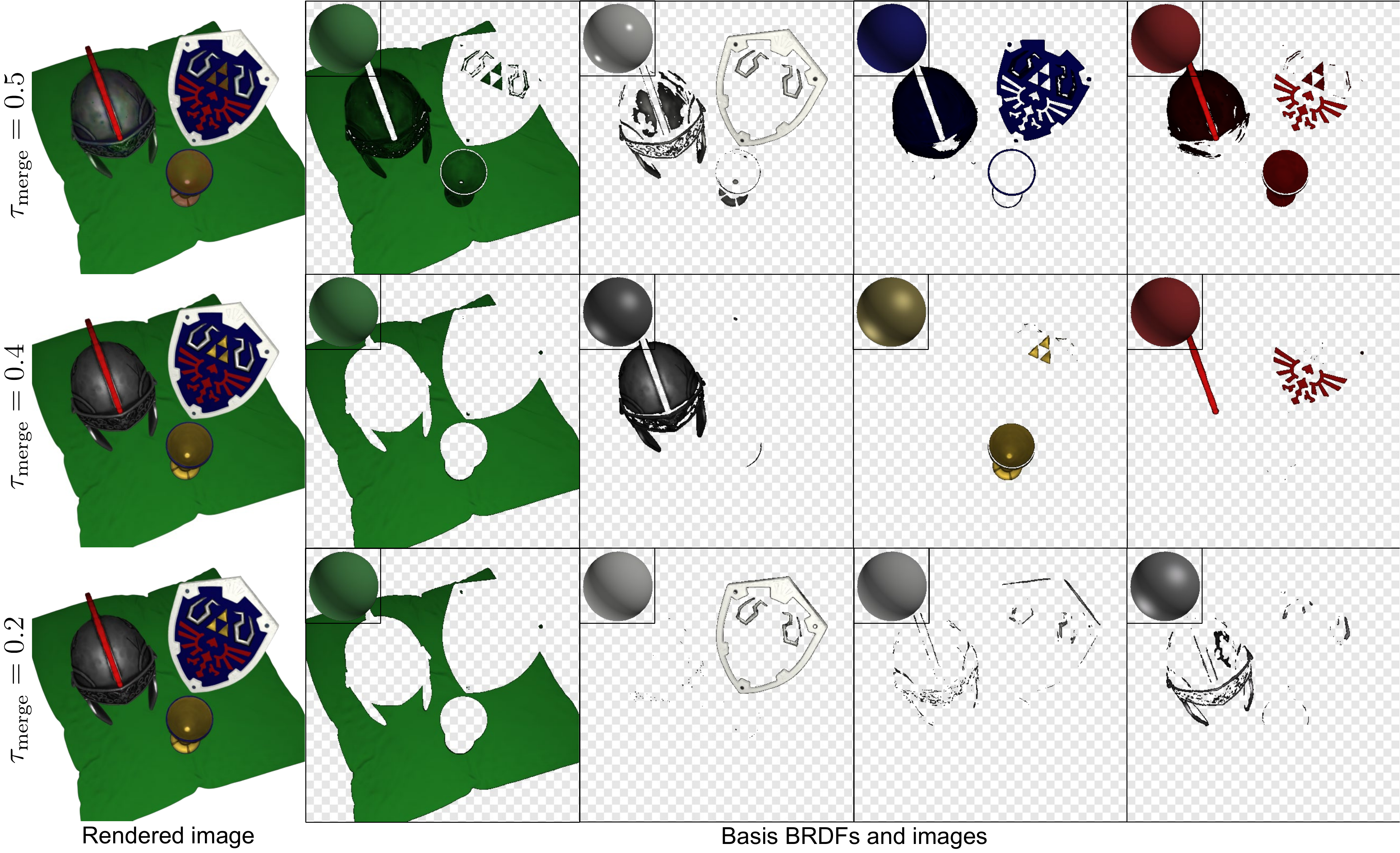}
    \caption{\textbf{Importance of merge threshold.} Merge threshold $\tau_{\text{merge}}$ directly affects the interpretability and scalability of basis BRDFs. If $\tau_{\text{merge}}$ is too large, distinguishable basis BRDFs are merged excessively. If $\tau_{\text{merge}}$ is too small, similar basis BRDFs are not merged. We utilize $\tau_{\text{merge}}= 0.4$ for this scene.}
\label{fig:ablation_merge} 
\end{figure}

Figure~\ref{fig:ablation_merge} shows the result of optimized basis BRDFs and images depending on the merge threshold $\tau_{\text{merge}}$. If we set a large value for the threshold $\tau_{\text{merge}}$, basis BRDFs are not merged when they are non-intuitive with duplicates. If we set a small value for the threshold $\tau_{\text{merge}}$, basis BRDFs are merged when they are not similar. We find appropriate merge threshold $\tau_{\text{merge}}$ for each complex scene to estimate optimal basis BRDFs and weight images.

\subsection{Thresholds for Basis BRDF Removal}
\begin{figure}[h]
    \centering
    \includegraphics[width=0.95\linewidth]{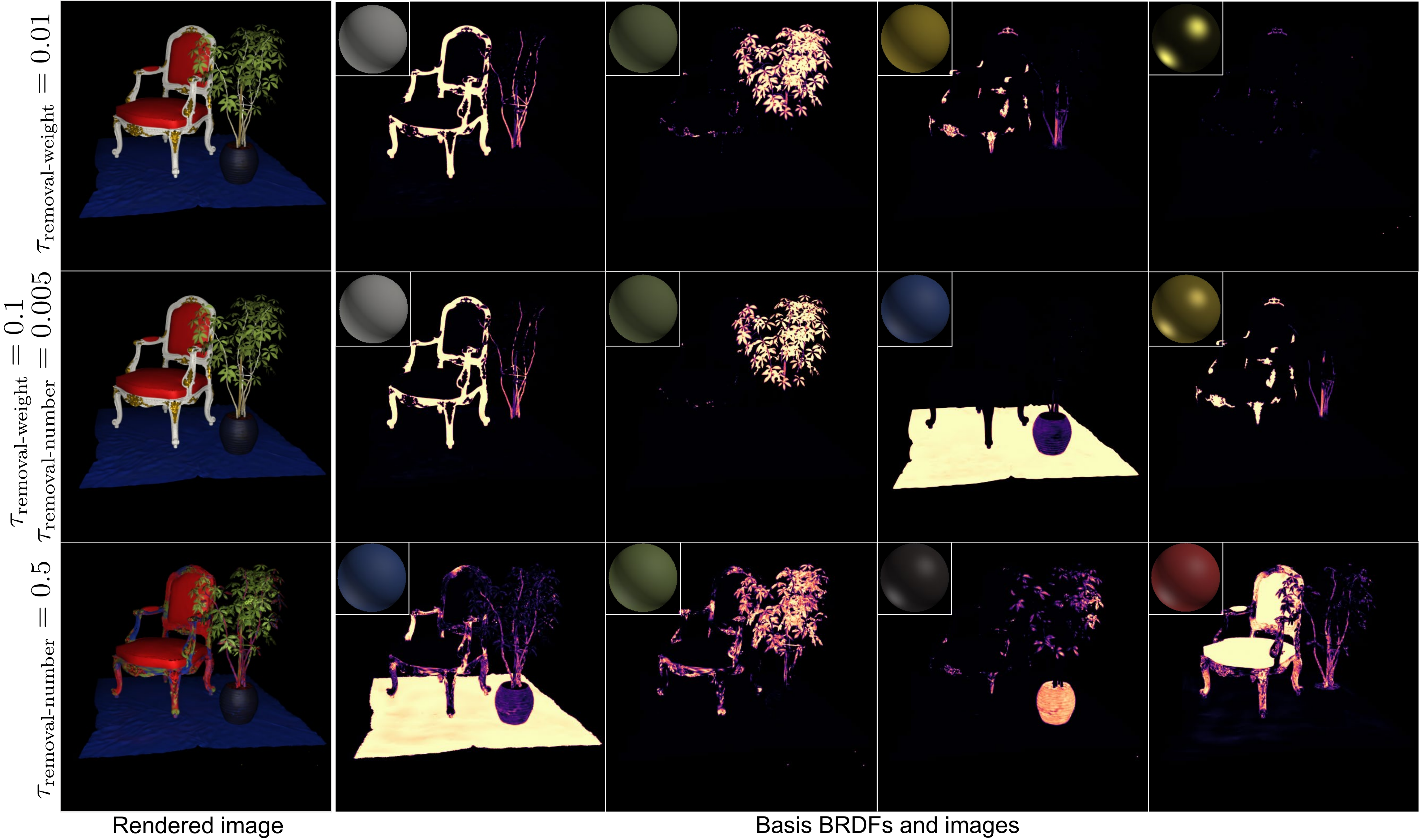}
    \caption{\textbf{Importance of removal thresholds.} Removal thresholds $\tau_{\text{removal-weight}}$ and $\tau_{\text{removal-weight}}$ determine which unnecessary basis BRDF to be removed. With small $\tau_{\text{removal-weight}}$, unnecessary basis BRDFs are not removed, resulting in non-interpretable basis BRDFs. If $\tau_{\text{removal-number}}$ is too big, necessary basis BRDFs are removed, and reconstruction fails. We utilize $\tau_{\text{removal-weight}}= 0.1$ and $\tau_{\text{removal-number}}= 0.005$ for this scene.}
\label{fig:ablation_removal} 
\vspace{-3mm}
\end{figure}

Figure~\ref{fig:ablation_removal} shows the result of optimized basis BRDFs and weight images depending on the removal thresholds $\tau_{\text{removal-weight}}$ and $\tau_{\text{removal-number}}$. $\tau_{\text{removal-weight}}$ determines whether weight image pixels are contributing to the reconstruction. $\tau_{\text{removal-number}}$ determines whether basis BRDFs are significant depending on the portion between valid weight pixels and total pixels. If $\tau_{\text{removal-weight}}$ is too small, many pixel are considered valid with non-interpretable basis BRDFs. If $\tau_{\text{removal-number}}$ is too big, necessary basis BRDFs are removed.

\section{Additional Discussions}
\subsection{Shadow Computation by Occlusion}
Our method utilizes captured images with multi-view flash photography to collect sufficient light-angular samples and neglect shadow by occlusion, which does not exist when light and camera are co-located. We achieve high-fidelity reconstruction of geometry and spatially-varying BRDFs even on real-world photometric dataset captured by a mobile phone with the flash light.
However, our method can be further improved with shadow computation by rendering a shadow image~\cite{bi2024rgs}. By multiplying a shadow image and a rendered image, we can render a photo-realistic image with shadow handling.

\subsection{Global Illumination}
We only consider direct illumination from a point light source, neglecting global illumination. Global illumination may be handled by employing residual MLPs or ray tracing. This would enable to handle indirect illumination and inter-reflection between specular objects, recovering accurate geometry and spatially varying BRDFs.


\section{Additional Results}
\subsection{Novel-view Relighting Results}
We compare our method with IRON~\cite{iron}, DPIR~\cite{chung2024differentiable}, GS\textsuperscript{3}~\cite{bi2024rgs} in quantitative manner. Table~\ref{tab:relighting_table} demonstrates that we achieve the highest accuracy with surface normal estimation and the second highest accuracy with novel view relighting. GS\textsuperscript{3} shows high-quality novel relighting results by employing residual networks with inaccurate geometry. Our method achieves both accurate geometry and interpretable basis BRDF reconstruction that enable high-fidelity novel view relighting and intuitive scene editing. 
\begin{table}[h]
\centering
\resizebox{0.4\linewidth}{!}{
    \begin{tabular}{c|ccc|c}
    \toprule[1pt]
            &\multicolumn{4}{c}{Synthetic photometric dataset} \\ \hline
    Method &PSNR $\uparrow$&SSIM $\uparrow$&LPIPS $\downarrow$   &MAE $\downarrow$ \\
    \hline
    IRON    &24.09	&0.8714	&0.0924	&15.71 \\
    DPIR    &26.87	&0.954	&0.0594	&17.04 \\
    GS\textsuperscript{3}   &\textbf{38.93}	&\textbf{0.9944}	&\textbf{0.0038}	&\underline{15.28} \\ \hline
    Ours   & \underline{31.78} & \underline{0.9754} & \underline{0.0234} & \textbf{9.81} \\
    \bottomrule[1pt]
    \end{tabular}}
    \caption{\textbf{Comparison between our method and other baselines.} We achieve high-fidelity novel-view relighting results with accurate geometry reconstruction.}
\label{tab:relighting_table}
\vspace{-3mm}
\end{table}

\subsection{Point Relighting Results}
\begin{figure}[h]
    \centering
    \includegraphics[width=0.95\linewidth]{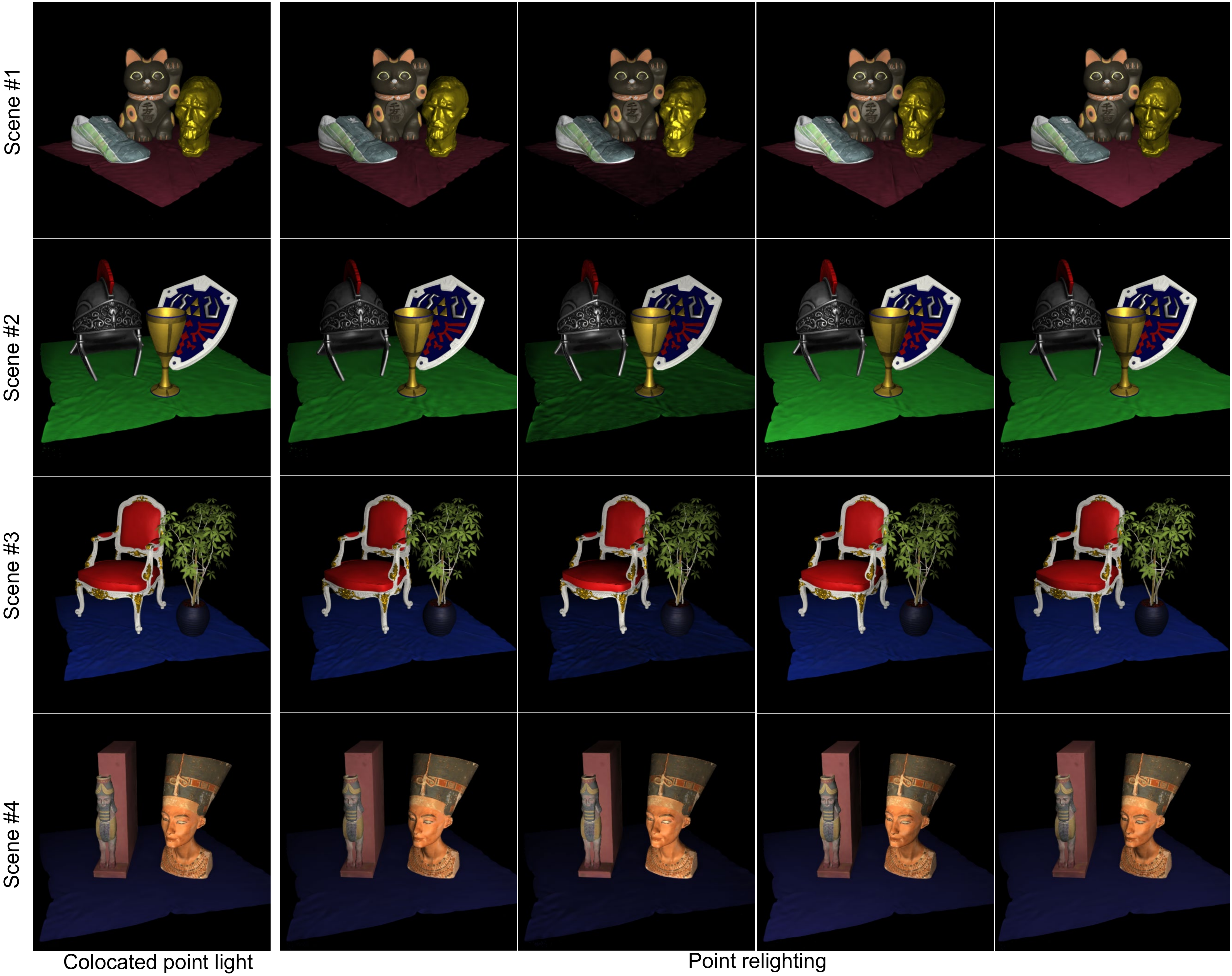}
    \caption{\textbf{Relighting with point light sources.} We render 4 complex scenes with a moving point light source.}
\label{fig:point_relighting_supp} 
\vspace{-3mm}
\end{figure}

We achieve realistic relighting results with a moving point light source as shown in Figure~\ref{fig:point_relighting_supp}. 

\subsection{Reflectance Editing Results}
\begin{figure}[h]
    \centering
    \includegraphics[width=0.95\linewidth]{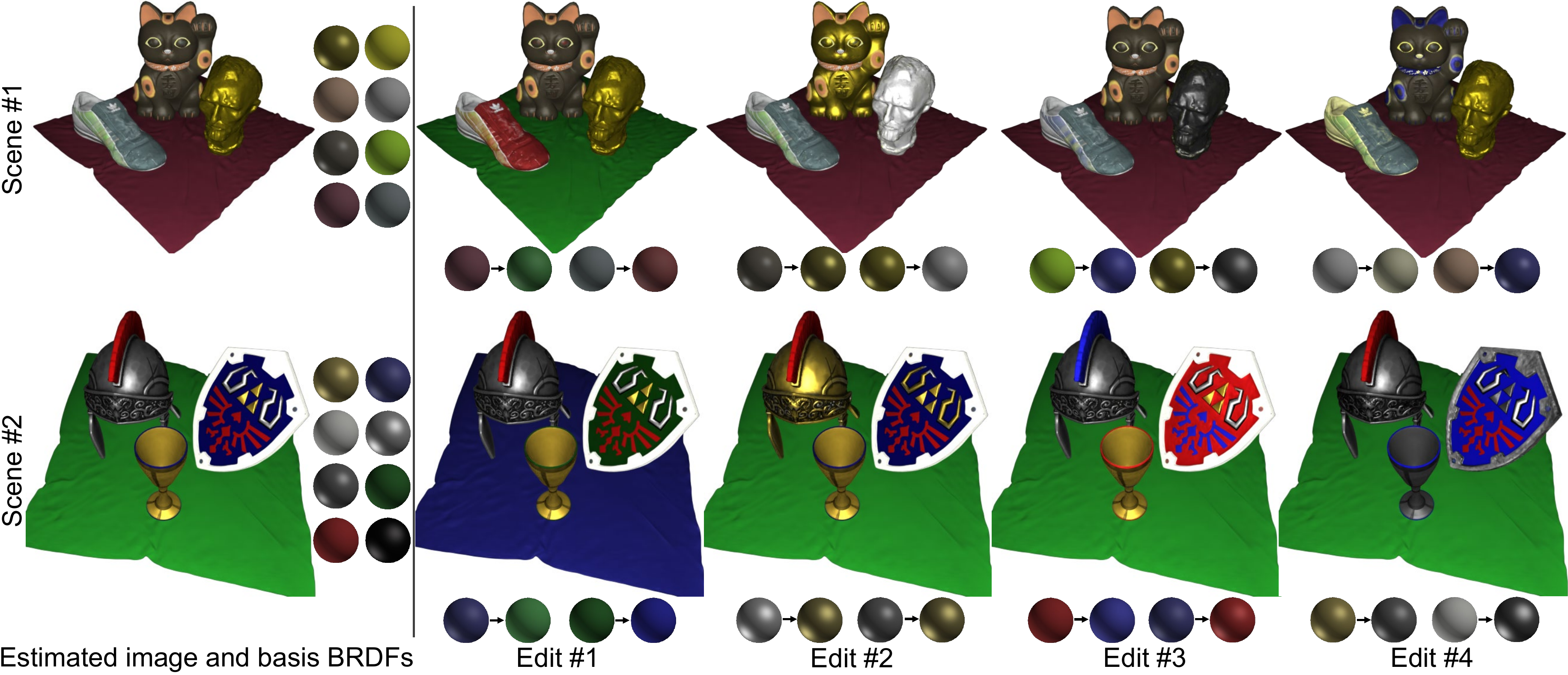}
    \caption{\textbf{Intuitive reflectance editing.} We efficiently edit reflectance by modifying parameters of basis BRDFs.}
\label{fig:reflectance_editing_supp} 
\vspace{-3mm}
\end{figure}

We achieve intuitive reflectance editing with interpretable basis BRDFs in Figure~\ref{fig:reflectance_editing_supp}. 

\subsection{Additional Results with Synthetic Photometric Dataset}
\begin{figure}[h]
    \centering
    \includegraphics[width=1.0\linewidth]{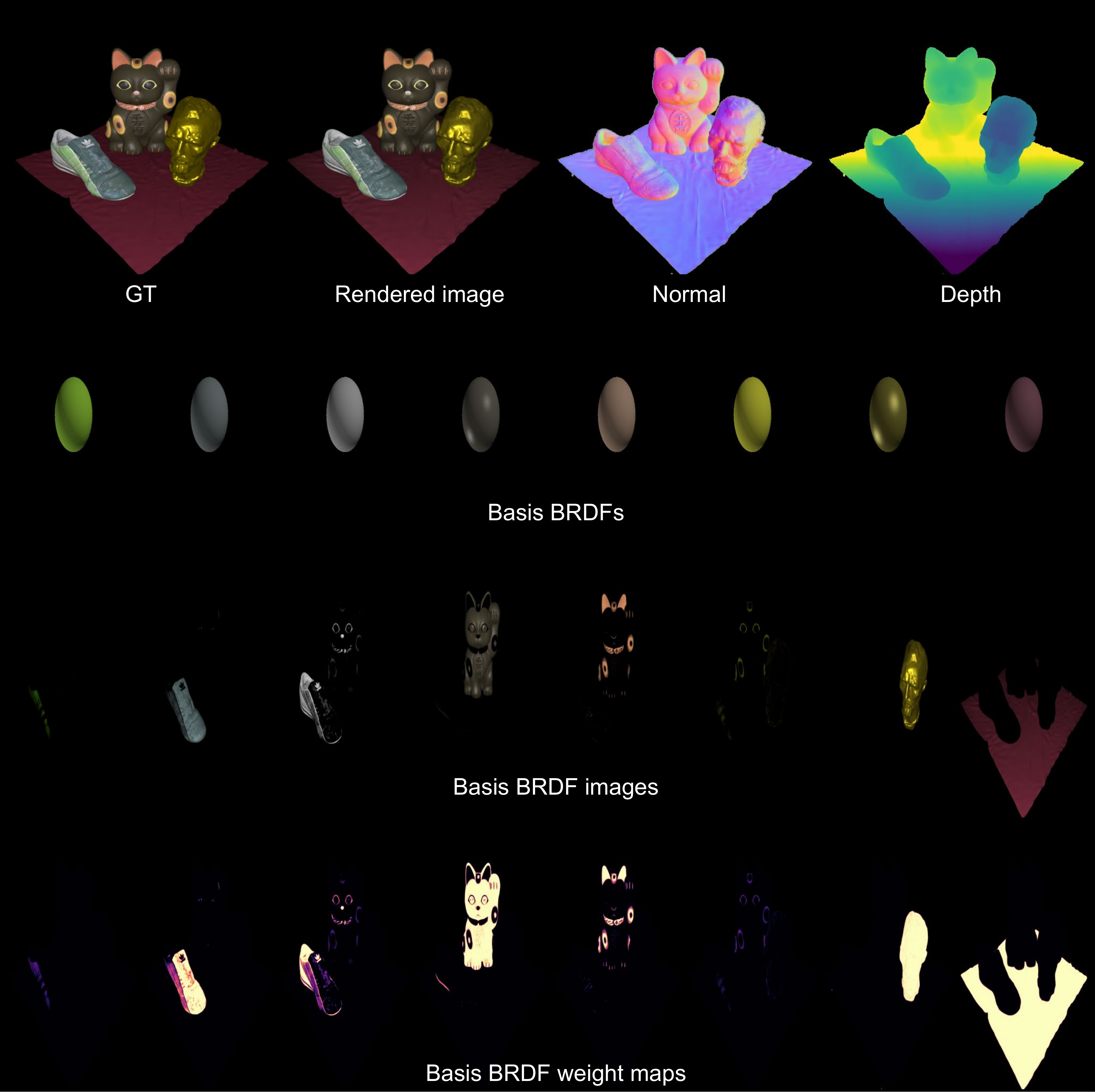}
    \caption{\textbf{Reconstruction results of scene $\text{\#} 1$ on the synthetic photometric dataset.}}
\label{fig:supple_synthetic1} 
\end{figure}

\begin{figure}[h]
    \centering
    \includegraphics[width=1.0\linewidth]{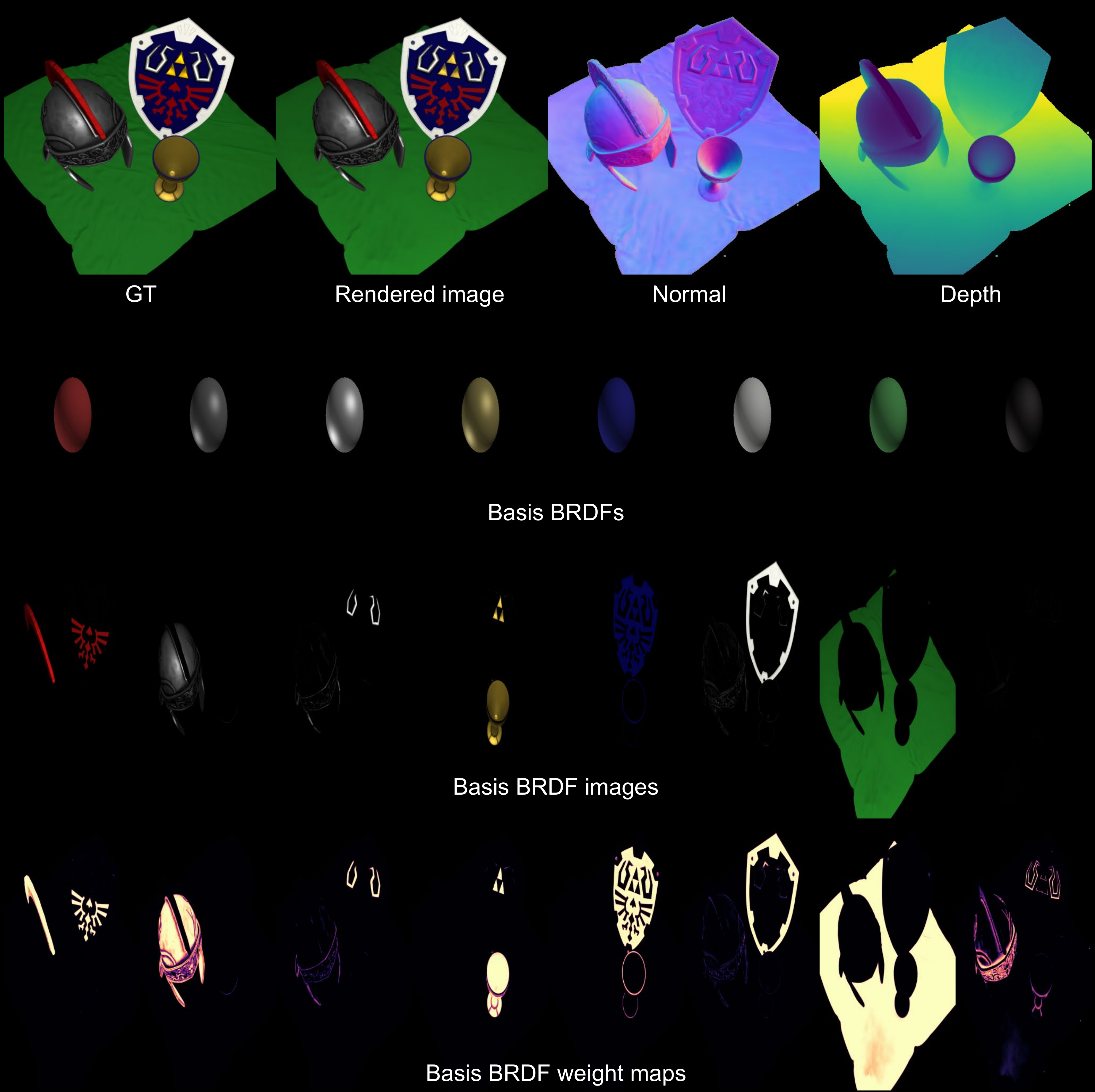}
    \caption{\textbf{Reconstruction results of scene $\text{\#} 2$ on the synthetic photometric dataset.}}
\label{fig:supple_synthetic2} 
\end{figure}

\begin{figure}[h]
    \centering
    \includegraphics[width=1.0\linewidth]{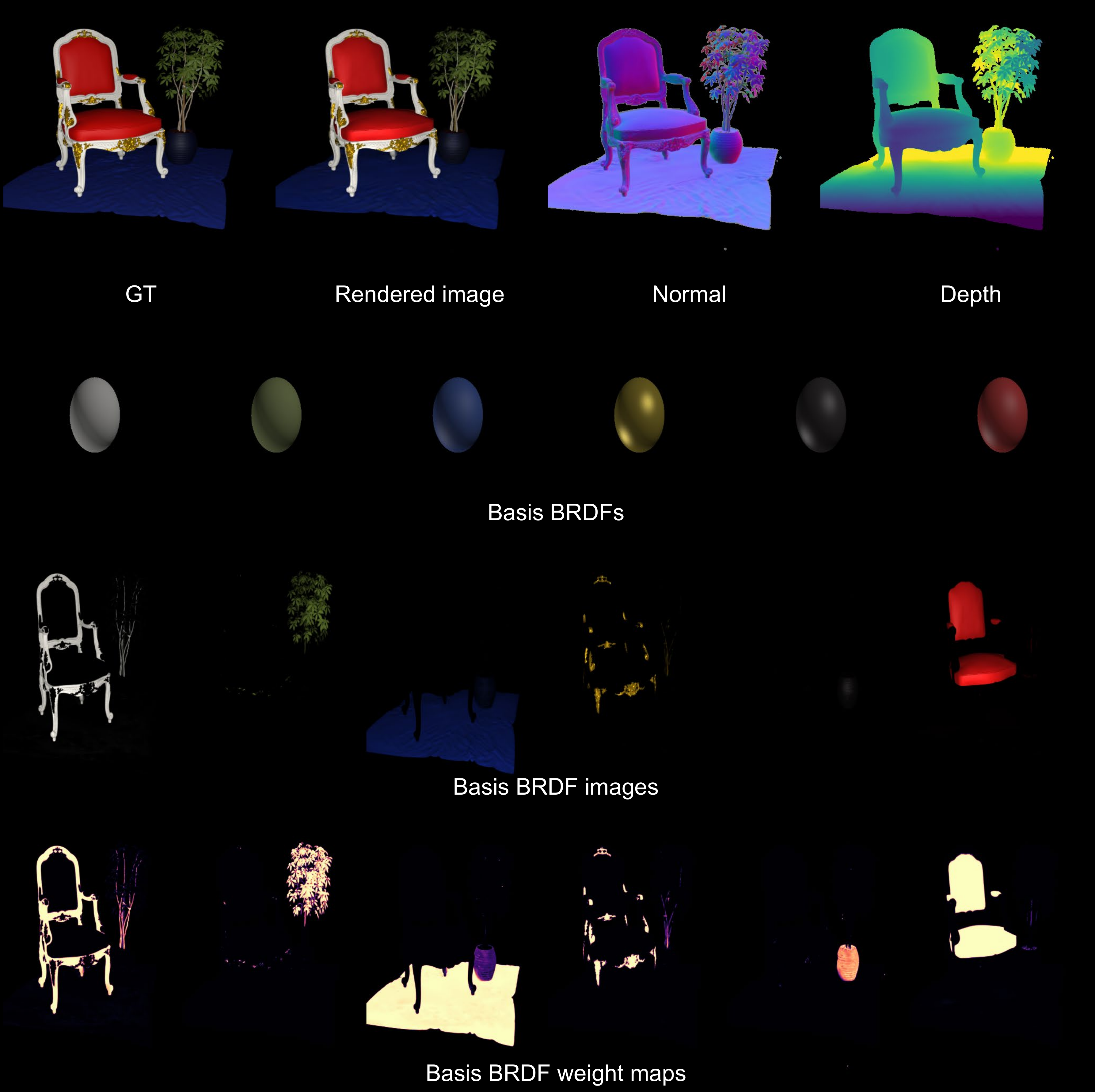}
    \caption{\textbf{Reconstruction results of scene $\text{\#} 3$ on the synthetic photometric dataset.}}
\label{fig:supple_synthetic3} 
\end{figure}

\begin{figure}[h]
    \centering
    \includegraphics[width=1.0\linewidth]{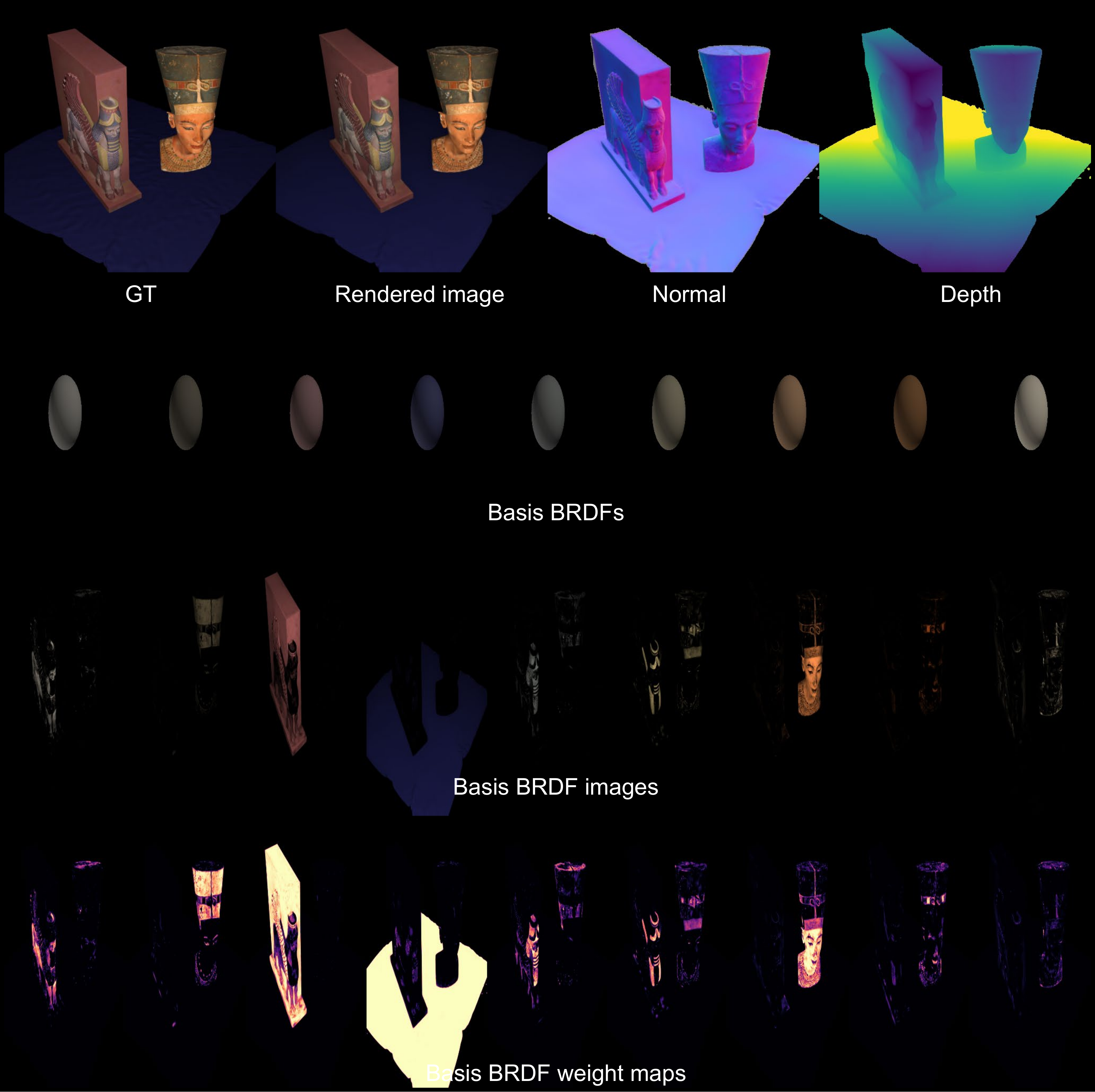}
    \caption{\textbf{Reconstruction results of scene $\text{\#} 4$ on the synthetic photometric dataset.}}
\label{fig:supple_synthetic4} 
\end{figure}

Figure~\ref{fig:supple_synthetic1}, Figure~\ref{fig:supple_synthetic2}, Figure~\ref{fig:supple_synthetic3} and Figure~\ref{fig:supple_synthetic4} show visualizations of 4 complex scenes in a synthetic photometric dataset. Our method reconstructs accurate geometry and interpretable basis BRDFs. We provide the visualization of a rendered image, estimated normal, depth map, basis BRDFs, basis BRDF images, and basis BRDF weight images. It demonstrates that our method successfully recovers complex geometry and spatially-varying BRDFs with interpretable basis BRDFs. Weight images are spatially separated which enables intuitive scene editing.

\clearpage

\clearpage

\clearpage

\clearpage

\subsection{Additional Results with Real-world Photometric Dataset}
\begin{figure}[h]
    \centering
    \includegraphics[width=1.0\linewidth]{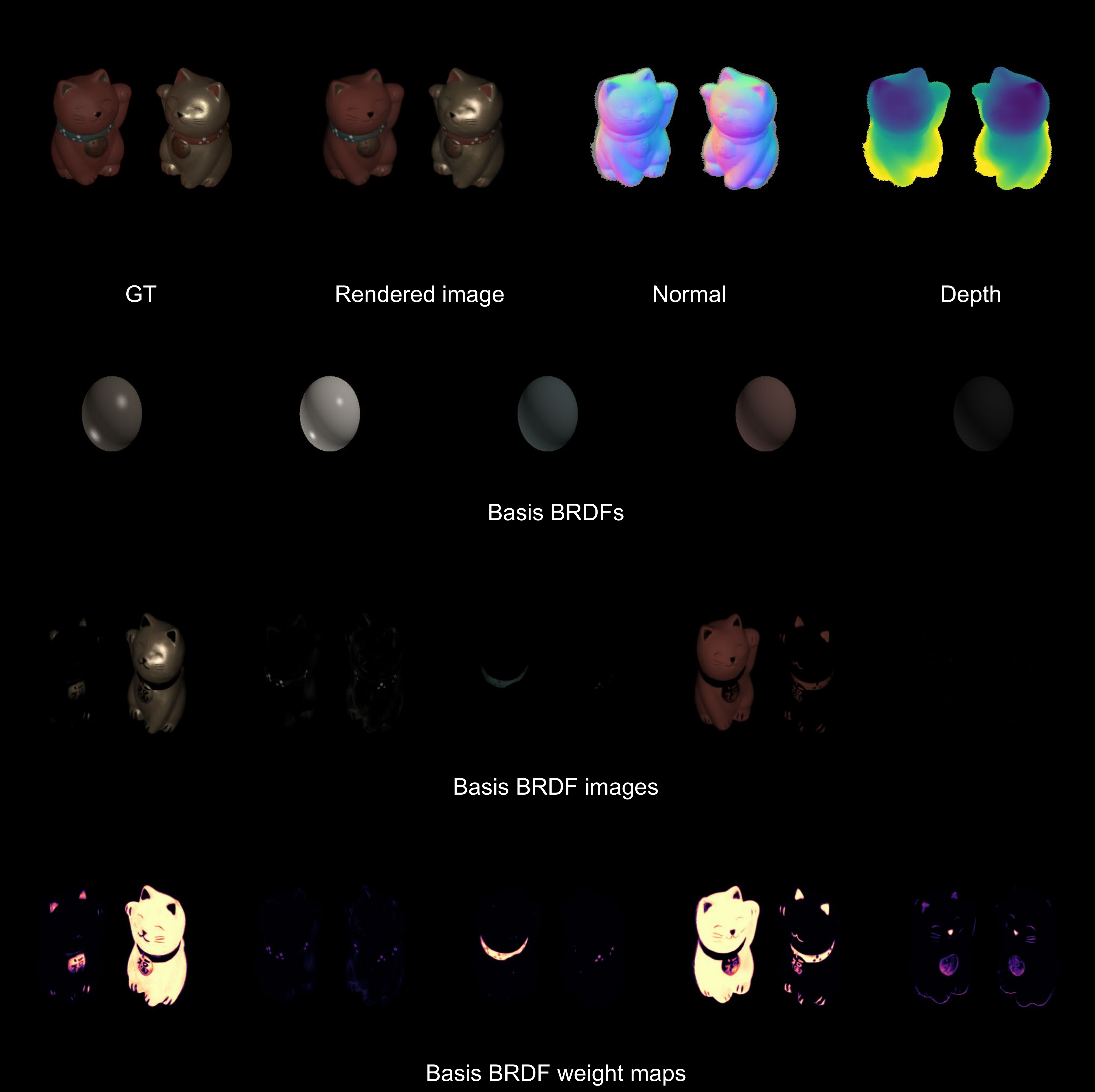}
    \caption{\textbf{Reconstruction results on the real-world photometric dataset.}}
\label{fig:supple_real} 
\end{figure}

Figure~\ref{fig:supple_real} shows a visualization of a complex scene in a real-world photometric dataset. Our method achieves accurate reconstruction of geometry and interpretable basis BRDFs from multi-view flash images.

\clearpage

{\small
\bibliographystyle{ieee_fullname}
\bibliography{references}
}

\end{CJK}